\documentclass[draftcls,onecolumn]{IEEEtran}

\usepackage{graphicx}
\graphicspath{{figuresEPS/}} 
\usepackage{subfigure}

\usepackage{amsfonts}

\usepackage[cmex10]{amsmath}
\newcommand{\Hm}{\bf{H}}
\newcommand{\Dm}{\bf{D}}
\newcommand{\Km}{\bf{K}}
\newcommand{\Lam}{\boldsymbol{\Lambda}}
\newcommand{\Sigm}{\boldsymbol{\Sigma}}
\ifCLASSINFOpdf
\else
\fi

\begin{document}
%
\title{A Parametric Level Set Approach to Simultaneous Object Identification and Background Reconstruction for Dual Energy Computed Tomography }
%
%
%

\author{Oguz~Semerci and Eric~L.~Miller,~\IEEEmembership{Senior Member,~IEEE}

\thanks{O. Semerci and E.L. Miller are with the Department
of Electrical and Computer Engineering, Tufts University, Medford,
MA, 02155 USA e-mail: elmiller@ece.tufts.edu.}
}

%
%


\markboth{SUBMITTED TO THE IEEE TRANSACTIONS ON IMAGE PROCESSING, 2011}%
{Shell \MakeLowercase{\textit{et al.}}: Bare Demo of IEEEtran.cls for Journals}

%



\maketitle

\begin{abstract}
Dual energy computerized tomography has gained great interest because of its ability to characterize the chemical composition of a material rather than simply providing relative attenuation images as in conventional tomography. The purpose of this paper is to introduce a novel polychromatic dual energy processing algorithm with an emphasis on detection and characterization of piecewise constant objects embedded in an unknown, cluttered background. Physical properties of the objects, specifically the Compton scattering and photoelectric absorption coefficients, are assumed to be known with some level of uncertainty. Our approach is based on a level-set representation of the characteristic function of the object and encompasses a number of regularization techniques for addressing both the prior information we have concerning the physical properties of the object as well as fundamental, physics-based limitations  associated with our ability to jointly recover the Compton scattering and photoelectric absorption properties of the scene. In the absence of an object with appropriate physical properties, our approach returns a null characteristic function and thus can be viewed as simultaneously solving the detection and characterization problems. Unlike the vast majority of methods which define the level set function non-parametrically, i.e., as a dense set of pixel values), we define our level set parametrically via radial basis functions (RBF's) and employ a Gauss-Newton type algorithm for cost minimization. Numerical results show that the algorithm successfully detects objects of interest, finds their shape and location, and gives a adequate reconstruction of the background.
\end{abstract}

\begin{IEEEkeywords}
Computed tomography, dual-energy, polychromatic spectrum, parametric level set, inverse problems, iterative reconstruction
\end{IEEEkeywords}

\IEEEpeerreviewmaketitle

\section{Introduction}
%
%
%
%

\IEEEPARstart{A}{ conventional} computed tomography (CT) imaging system provides a reconstruction of the
linear attenuation coefficient distribution of an object under
investigation. Dual energy techniques, however, allow for more
detailed chemical characterization of the material using measurements
from two distinct X-ray spectra.  These methods have been applied in a
range of application areas including non-destructive material
evaluation \cite{engler1990review}, medical imaging \cite{johnson2007material}, cardiac and coronary
imaging \cite{achenbach2008dual}, bone densitometry \cite{kroger1992bone} and airport/seaport
security \cite{ying}.

The dual energy idea was initially proposed by Alvarez and Macovsky
\cite{alvarez} who modeled the total attenuation of X-rays as a linear
combination photoelectric absorption and Compton scattering
coefficients with corresponding empirical energy dependent
basis functions. In detail, a cubic polynomial approximation of the
polychromatic measurement models was used to estimate sinograms of
photoelectric and Compton components. Once sinograms were estimated,
the filtered back projection (FBP) algorithm was applied to obtain
photoelectric and Compton coefficient reconstructions. The polynomial
approximation, however, was found to be accurate only for test subjects
whose properties were ``close enough'' to data subjects used to obtain
the model coefficients. Also the sinogram estimation step amplified
the noise leading to erroneous reconstructions especially for the
photoelectric component \cite{alvarez}. Later, basis material decomposition
methods, where attenuation coefficients of two or more materials
constitute the basis set for the total attenuation, were proposed
\cite{lehmann,kalender}. For these methods, the reconstruction problem was reduced to
estimation of the space-varying weight of each basis material.  This
approach is accurate when the properties in the materials of interest
are, in a sense, ``spanned'' by the spectral characteristics of the
basis material \cite{zhang}. Optimal selection of the basis set is
generally application dependent with a variety of approaches having
been explored in the literature to date \cite{goh1997correction,williamson2006two,heismann2}. A review of the accuracy of these methods can be found in \cite{heismann2}. The methods where the estimation of sinograms of basis components is followed by FBP reconstructions of the corresponding images are referred as prereconstruction methods. Alternatively in the case of postreconstruction type of methods, filtered back projection is used to form separate attenuation images from the high and low energy projection data which are then mapped into basis material images \cite{joseph1978method,engler1990review}. It is well-known that both of these conventional approaches to the dual energy problem have poor noise properties and are prone to inaccuracies in the measurement data as they use FBP \cite{sukovic,kalender2002algorithm,de2001iterative}.


In addition to reconstruction methods based on FBP, significant work
has been directed toward the use of iterative image formation schemes
for dual energy CT.  Such methods provide the opportunity to more
precisely take into account the nonlinear relationship between the
data and the material properties than is afforded by FBP-based
algorithms.  Moreover, iterative techniques can be tailored to the
underlying Poisson nature of the observations and provide a natural
mechanism for incorporating prior information into the image formation
process \cite{kaipio2005statistical}. Within the context of dual energy CT, initial work
on iterative methods focused on using the algebraic reconstruction
technique (ART) and basis material decomposition for iterative beam
hardening correction \cite{markham1993element,michael1992tissue}. More recently, iterative methods based
on maximum likelihood and maximum \emph{a posteriori} (MAP)
statistical principles have been developed for monochromatic and
polychromatic dual energy reconstruction problems \cite{sukovic, de2001iterative, fessler&sukovic} arising primarily in the medical imaging
domain.  Such methods have proven to be more accurate than previously
proposed iterative and FBP based approaches especially in low photon
count (i.e., low SNR) scenarios.

While medical applications have motivated a significant majority of
algorithm development for dual energy CT, in recent years, there has
been growing interest in the application of this imaging modality to
problems in airport and seaport security; specifically the screening
of checked luggage \cite{ying,ying2007dual,butler2002rethinking,singh2003explosives}.  Unlike the medical imaging problem, the range of
materials encountered in baggage screening is quite broad.  As such,
it remains an open question as to whether and how material basis-type
decompositions of the attenuation coefficient can be applied in this
context.  Indeed, the state-of-the art here is represented by the work
of Ying \emph{et al.} who consider the recovery of the Compton and
photoelectric coefficients directly in \cite{ying}. Following a similar
approach to \cite{alvarez}, Ying \emph{et al.}  employ a prereconstruction imaging
scheme in which the sinograms for the photoelectric and Compton
coefficients are obtained via the solution of a non-linear constrained
optimization problem. As was the case in \cite{alvarez}, high fidelity
recovery of the photoelectric coefficient proved challenging due to
the domination of the Compton effect relative to the photoelectric for
this class of problems.  Roughly speaking the signal to noise ratio
associated with the photoelectric component was far smaller than that
of the Compton making stable recovery of the former difficult.

Consideration of these issues associated with the use of dual energy
CT for luggage screening provides the impetus for the work in this
paper.  To date, the algorithms developed for the screening problem
are rather similar to those developed for medical applications despite
the fact that the fundamental objectives of these two problems are
somewhat different.  In the medical case, the processing goal is the
creation of high resolution images capable of supporting accurate
diagnoses.  For the security problem however, while one may still
desire a good ``picture,'' an additional goal is the determination as
to whether a given object contains illicit materials and if so, how
are they distributed.  Such materials of interest will be embedded in
an unknown and generally quite inhomogeneous background; however,
prior information exists concerning the types of materials that may be
of interest.  Our approach to bringing this and related prior
information to bear on the image formation process has not, to the
best of our knowledge, been considered.

Motivated by these observation, in this paper we propose a variational
approach to image formation adapted to the problem of detecting and
characterizing ``anomalies'' in multiple physical parameters located
in a cluttered and unknown environment.  The contributions of this
work are as follows.  First we propose a new, hybrid model and
associated inversion methods for the determination of the geometry and
contrast associated with these anomalies along with a suitable
pixel-like representation of the unknown background.  Should an
anomaly not be present, images of the physical properties are
provided.  In the case where a anomalies do exist, in addition to the
images, we obtain an explicit representation of the shape, location,
and contrast of the objects.  Thus, the method we propose here
simultaneously addresses both the detection and characterization
problems.

Second, we develop a new regularization scheme for problems in which
there is a mismatch in the sensitivity of the data to the various
properties being imaged.  As noted above, for the dual energy CT
problem, the Compton scatter effects tends to dominate the
photoelectric effect. Such challenges however are not limited to CT.
For example, when using diffuse optical tomography for medical imaging
problems, we have found sensitivity mismatch when attempting to
recover multiple chromophore concentration profiles \cite{larusson2009hyperspectral}.
More generally, the ideas considered here are of potential use to a much
broader range of so-called joint inversion problems where a collection
of heterogeneous sensors (acoustic, electromagnetic, optical,
mechanical, hydrological) are tasked with developing a single
``picture'' of a given region of space.  With each modality sensitive
to its own constitutive properties of the medium, the modeling and
regularization ideas considered here represents a step in fusing
information in such scenarios. An example for such an inversion scheme arises in the geophysical imaging context \cite{gallardo2003characterization} where a joint inversion of DC resistivity and seismic refraction is performed to model subsurface profiles where cross product of the gradients of the reconstructed parameters are used to ensure that the profiles provided by the inversion of each parameter are in total agreement.

While the we present here approach is \emph{potentially} of broad use, in this paper
the \emph{specific} concern is its application to the dual energy CT
for luggage screening problem.  Here, we view the work as a contribution to
the use of iterative inversion schemes for this class of problems.
Like the MAP approach, image formation is cast as the solution to a
variational problem comprised of terms reflecting a level of fidelity
to the data and prior information.  In this context, the model we
propose and the structure of the prior information are new to the dual
energy CT literature.  In a bit more detail, as was the case in
\cite{ying}, we consider here the recovery of the space-varying structure
of the photoelectric and Compton scattering parameters.\footnote{While
  a basis-material decomposition could be employed, as discussed
  previously, it is not clear at the present time which materials are
  most suitable for the screening problem.  Moreover, one contribution of the work here is a new approach to
  stabilize the recovery of the photoelectric coefficient.}  We model
these quantities in a hybrid manner as the superposition of a
parametrically defined anomaly (should it exists) and a non-parametric
background.

The anomaly model is comprised of a geometric component that is common
in the two images and contrast parameters that are specific to the
photoelectic and Compton images.  In this paper, we employ a newly
developed parametric level set based representation for the anomalies.
Classical level-set methods was developed for modeling propagation of
curves by Osher and Sethian \cite{osher1} and have been widely used in image
processing applications \cite{chan,samson,tsai} as well as for a variety of inverse
problems \cite{santosa,lesselier,miled,feng2003curve,kolehmainen2008limited}.  They provide a topologically flexible shape-based
formulation, elegantly representing multiple objects with complicated
geometries.  The number of unknowns associated with such a model
however is equal to the number of pixels in an image.  As such they
can be difficult to use when solving inverse problems \cite{aghasi2010parametric}.
Recently, level set methods have been defined based on low order basis
functions expansions for the level set function \cite{kilmer2004cortical}. In the
context of image segmentation and topology estimation gradient
decent-type methods have been developed for determining the unknown
expansion coefficients describing the level set function \cite{r1992alternatives,bernard,pingen,wang2003level,gelas2007compactly}. An
alternative parametric level set model specifically adapted to solving
inverse problems was considered in \cite{aghasi2010parametric}.  As discussed in \cite{aghasi2010parametric},
this model is especially attractive as there is no need for reinitialization or narrow-banding; all
issues frequently encountered with curve evolution methods for inverse
problems. It suffices to incorporate a regularization term that penalizes anomalies with large areas. Moreover, because the order of the model is low (on the
order of 10's of unknowns) quasi-Newton methods become quite feasible
for the estimation of the parameters.

With these advantages, it is this model that forms the basis for the
work here. We apply the parametric level set idea to the dual energy CT problem where detection of the size number and location of objects plays a crucial role. As such we we demonstrate empirically that the detection and localization of these objects becomes possible even if the photoelectric image background reconstruction is not very accurate or a FBP based approach would fail to provide reasonable results.


In addition to this approach for describing the shapes of the anomalies, the second component of our anomaly model is the contrast of these
object in both the Compton and photoelectric images.  Here we exploit
the fact that, while the nominal background may be quite variable, for
applications such as the one motivating this work, one is searching
for objects of specific composition.  In other words, the physical
properties (Compton scattering and photoelectric coefficients) of any
anomaly are known at least to within some degree of precision.  We
interpret this type of information as providing a set of allowable
values for the contrast of the object in the photoelectric - Compton
coefficient space.  By definition then, this set of values
\emph{cannot} be assumed by any pixels \emph{not} in the object; i.e.,
pixels in the background.  These two constraints are enforced
explicitly in the variational formulation of the image formation
problem.

While a parametric model is used to represent the anomaly, given the
lack of structure we assume for the background, a pixel-type approach
is used to model the nominal spatial structure of the photoelectric
and Compton images.  As noted previously, the physics of dual energy
CT are such that in noisy situations one can recover the Compton image
rather well but the photoelectric image tends to be far less accurate
\cite{ying,alvarez,heismann2}. In an attempt to improve this situation, here we take
advantage of the fact that the scene being imaged is the same for both
parameters.  More specifically, because physical objects are
represented by discontinuities in the constitutive properties of the
medium, one would expect that edges in both images should be highly
correlated.  Hence, we introduce a correlation based regularization
scheme enforcing structural similarity between the gradients of reconstructed images.

The remainder of this paper is organized as follows. In Section \ref{sec:poly}, we provide the polychromatic CT measurement model as well as the representation of total attenuation in terms of photoelectric effect and Compton scatter. In Section \ref{sec:pals} we describe the shape based model via a parametric  level set function and formulate the Compton and photoelectric images in terms of homogeneous objects of known properties and unknown background pixels. In Section \ref{sec:reconstruction_algorithm} the numerical strategy for the solution is described. We cast the inverse problem in an optimization framework in which the solution is the minimizer of a cost function which consists of a data mismatch term and regularization terms. To solve the optimization problem, an alternating minimization algorithm for the shape and background parameters is applied with Levenberg-Marquardt algorithm employed in each phase. Numerical examples are given are given in Section \ref{sec:numerical_examples}. Section \ref{sec:conclusions} consists of concluding remarks and thoughts about future goals. Finally in the Appendix, we consider a linearized and simplified version of the background reconstruction problem and derive lower bounds for the mean square errors (MSE) in photoelectric absorption and Compton scatter coefficient estimation. Showing the the bounds are significantly different, we give the reader the intuition about the difficulties that arise in photoelectric absorption coefficient reconstruction and explain the motivation that derived us to use the correlation based regularization.

\section{Polychromatic Computerized Tomography Formulation}
\label{sec:poly}

\begin{figure}
\centering
\includegraphics[width=7 cm]{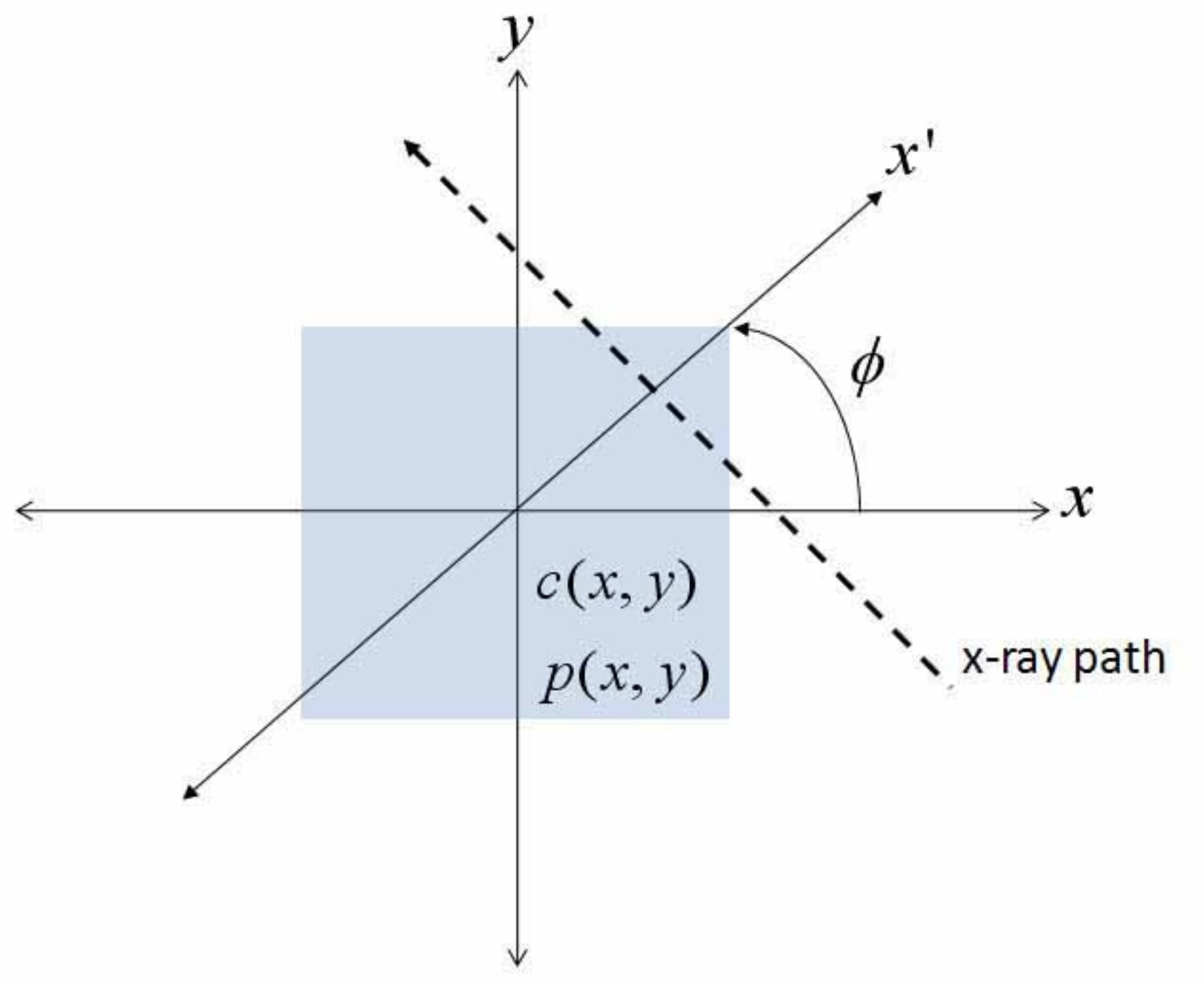}
\caption{Geometry of the Radon transform}
\label{fig:radon}
\end{figure}

Typical X-ray sources used in CT applications generate an energy spectra roughly between $20$KeV and $140$KeV \cite{beutel2000handbook}. In this energy range X-ray attenuation is dominated by Compton scatter and photoelectric absorption \cite{alvarez} each of which can be modeled as a product of an energy and material dependent terms as follows
\begin{equation}
\label{attenuation}
\mu(x,y,E) = c(x,y)f_{KN}(E)+p(x,y) f_p(E)
\end{equation}
where $\mu(x,y,E)$ is the total attenuation, $c(x,x)$ and $p(x,y)$ are the material dependent Compton scatter and photoelectric absorption coefficients respectively. The quantity $f_{KN}$ is the Klein-Nishina cross section for Compton scattering which is given as:
%
%
\begin{equation}
\label{klein-nishina}
\begin{split}
f_{KN}(\alpha) & = \frac{1+\alpha}{\alpha^2}\left[\frac{2(1+\alpha)}{1+2\alpha}-\frac{1}{\alpha}\ln(1+2\alpha)\right] \\
& \quad  +\frac{1}{2\alpha}\ln(1+2\alpha)-\frac{1+3\alpha}{(1+2\alpha)^2}
\end{split}
\end{equation}
where $\alpha=E/510.95$KeV.  Lastly, $f_p$ approximates the energy dependency of the Photoelectric absorption and given as
\begin{equation}
\label{fp}
f_b = E^{-3}.
\end{equation}
In polychromatic CT the measured quantities are logarithmic projections obtained at several measurement points defined formally as:
\begin{equation}
\label{projection}
P(\phi,x') = -\ln\frac{Y( \phi,x')} {Y_0}.
\end{equation}
Here $Y( \phi,x')$ is a Poisson random variable with mean
\begin{equation}
\begin{split}
\overline{Y}(\phi,x')& = \int S(E)\mathrm{exp}\bigl(-f_{KN}(E)A_c(\phi,x') \\
                     & \quad - f_{p}(E)A_p(\phi,x')\bigr)\mathrm{d}E + r(\phi,x')
\end{split}
\end{equation}
where $S(E)$ is X-ray spectrum, $\phi$ is the measurement angle, $x'$ is tilted version of $x$ axis by $\phi$,  $r(\phi,x')$ is the background signal caused by scatter and detector noise;  $A_c(\phi,x')$ and $A_p(\phi,x')$ are the Radon transforms for the x-ray path (see Fig. \ref{fig:radon}) of each measurement of Compton scatter and photoelectric absorption coefficients respectively and given as the following:
\begin{equation}
\label{Ac}
A_c(\phi,x')= \int c(x,y)\delta(x\cos\theta+y\sin\theta-x')\mathrm{d}x\mathrm{d}y,
\end{equation}
\begin{equation}
\label{Ap}
A_p(\phi,x')= \int p(x,y)\delta(x\cos\theta+y\sin\theta-x')\mathrm{d}x\mathrm{d}y.
\end{equation}
We call $c(x,y)$ the Compton and $p(x,y)$ the photoelectric images. The blank scan, $Y_0$, is assumed to be constant for all $(\phi,x')$ and given as
\begin{equation}
Y_0 = \int S(E)\,\mathrm{d}E.
\end{equation}
We define the reconstruction domain $\mathcal{D} \subset \mathbb{R}^2$ as a two dimensional rectangular region such that
\begin{equation}
\label{D}
\mathcal{D} = \{(x,y):0\leq x\leq l_x, 0\leq y\leq l_y \}.
\end{equation}
In order to obtain a discrete representation of (\ref{Ac}) and (\ref{Ap}) we assume that $c(x,y)$ and $p(x,y)$ are piecewise constant on each pixel of a regular grid with collection of $N_p$ pixels and  define the vectors $\mathbf{c}=[c_1,\dots,c_{N_p}]^T$ and  $\mathbf{p}=[p_1,\dots,p_{N_p}]^T$  to be lexicographically ordered collection of pixels values. Now, (\ref{Ac}) and (\ref{Ap}) can be rewritten in a matrix form using system matrix $\textbf{A} = \{ a_{ij} \}$  where  $a_{ij}$ represents the length of that segment of ray $i$ passing through pixel $j$.

In this work we assumed parallel beam projections performed for $N_{\theta}$ different angles between $0$ and $180$ degrees. For each angle, x-ray intensities on equally spaced $x'$ points are recorded to give $N_m$ measurements in total. The measurement is repeated for low and high energy source spectra to constitute the vector $\mathbf{m}^T = [ \mathbf{m}_L^T, \mathbf{m}_H^T ]$ which consists of $2N_m$ elements. Using the discrete grid and the system matrix $\mathbf{A}$, the $i$th measurement for low and high energy levels are written as follows:
\begin{equation}
\label{projectiondL}
[\mathbf{m}_L]_i  = -\ln\frac{[\mathbf{Y}_L]_i} {Y_{0,L}}
\end{equation}
and
\begin{equation}
\label{projectiondH}
[\mathbf{m}_H]_i  = -\ln\frac{[\mathbf{Y}_H]_i} {Y_{0,H}}.
\end{equation}
Similar to the continuous case $\mathbf{Y}_L$ and $\mathbf{Y}_H$ are vectors of Poisson random variables with means
\begin{equation}
\label{YdL}
\begin{split}
[\overline{\mathbf{Y}}_L]_i & = \int S_L(E)\mathrm{exp}\bigl(-f_{KN}(E)\mathbf{A}_{i*}\mathbf{c} \\
&\quad -f_{p}(E)\mathbf{A}_{i*}\mathbf{p}\bigr)\mathrm{d}E + r_{L,i}
\end{split}
\end{equation}
and
\begin{equation}
\label{YdH}
\begin{split}
[\overline{\mathbf{Y}}_H]_i & = \int S_H(E)\mathrm{exp}\bigl(-f_{KN}(E)\mathbf{A}_{i*}\mathbf{c} \\
&\quad -f_{p}(E)\mathbf{A}_{i*}\mathbf{p}\bigr)\mathrm{d}E + r_{H,i}
\end{split}
\end{equation}
\begin{figure}
\centering
\includegraphics[width=8 cm]{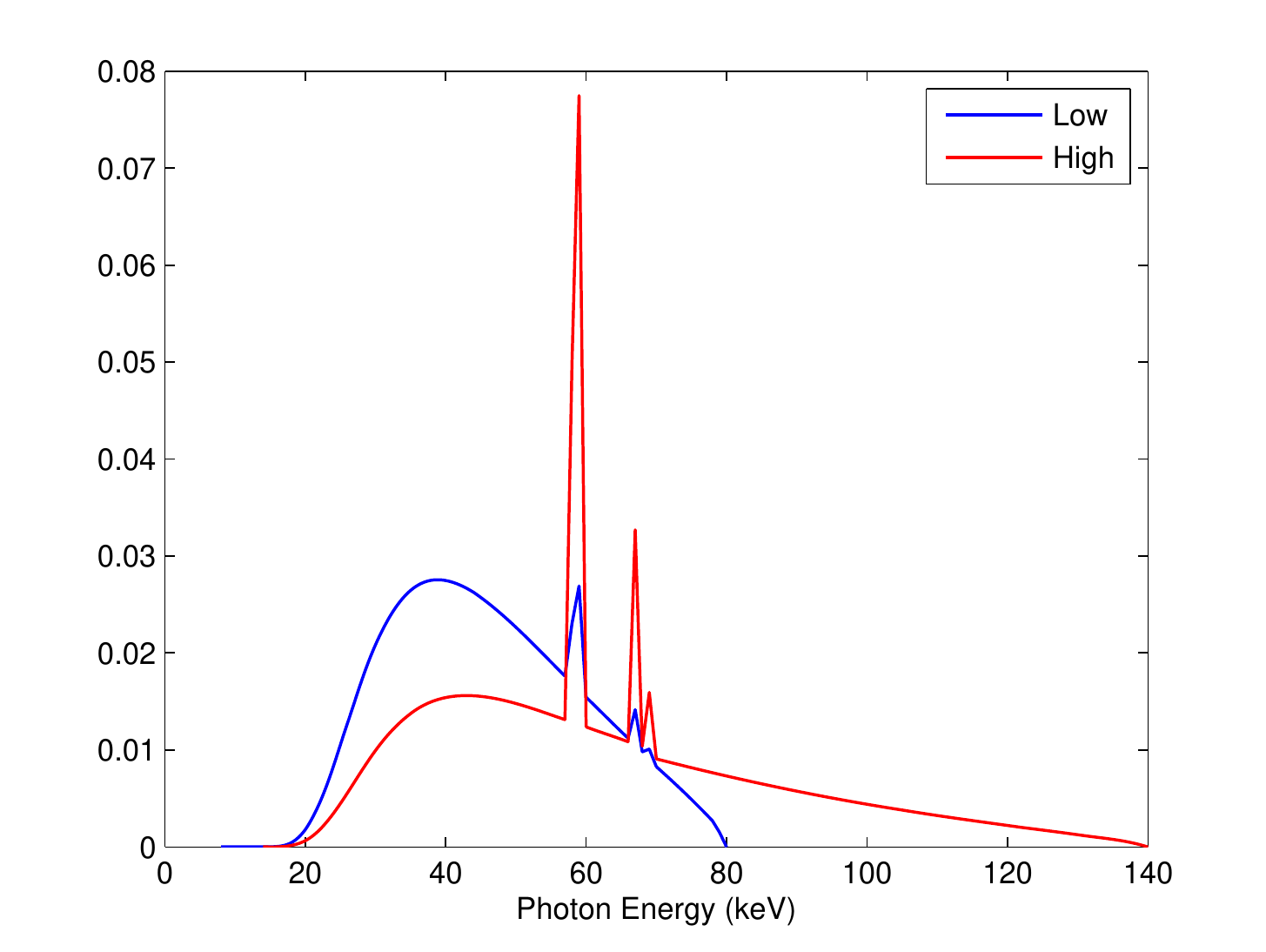}
\caption{Normalized X-ray Energy Spectra}
\label{fig:spectra}
\end{figure}
where $\mathbf{A}_{i*}$ is the i$^{\mbox{th}}$ row of $\mathbf{A}$, $S_L(E)$ and $S_H(E)$, corresponds to low energy and high energy X-ray spectra which are shown in Fig. \ref{fig:spectra}. In our work, these spectra with 1 keV energy bins were obtained using \textit{SpekCalc}, a free of charge software program which calculates x-ray spectra from tungsten anode tubes where specifications such as tube voltage and filtration thickness are assigned via a friendly user interface \cite{evans2009spekcalc,poludniowski2007calculation1,poludniowski2007calculation2}. Distinct energy levels are obtained by alternating the tube voltage between 80 keV and 140 keV. The total number of photons for $S_L(E)$ is $1.8 \times 10^6$ and for $S_H(E)$ is $3.6 \times 10^6$. The combination of scatter and detector readout noise contributions are denoted by $r_{L,i}$ and $r_{H,i}$. In practice these contributions are not known \textit{a priori} since they are related to object properties (shape and chemical composition) and scanner specification. However, scatter is a smoothly varying signal which can be estimated \cite{elbakri2003segmentation, hansen1997extraction} and often assumed known \cite{elbakri2003segmentation} or characterized by constant primary-to-scatter ratio \cite{joseph1982effects,de1999metal}. On the other hand, detector readout noise characteristics can be approximated by a Gaussion distribution \cite{tang2009performance}. Here, we assume $r_{L,i}$ and $r_{H,i}$ are additive white Gaussian (AWGN) and call them background noise. For a more advanced treatment of scatter see \cite{wiegert2007scattered}.

\section{The Parametric Level-Set Model and Representation of Compton and Photoelectric Images}
\label{sec:pals}


Compton scatter and photoelectric absorption coefficients are the main interests of our polychromatic CT problem and can be regarded as two different images to be reconstructed.  Each image is assumed to be formed by piecewise constant objects of interest on an unknown background. For a domain $\Omega \subset  \mathcal{D}$ which represents the support of the objects of interest, the characteristic function $\chi(x,y)$ is defined as
\begin{equation}
\label{chi}
\mathcal{\chi}(x,y)= \left \{ \begin{array}{cl}
1,& \mbox{  if}(x,y)\in\Omega \\
0,& \mbox{  if}(x,y)\in \mathcal{D}\backslash\Omega.  \\
\end{array} \right.
\end{equation}
Now, the Compton and photoelectric images can be written as
\begin{equation}
\label{cimage}
c(x,y)=\chi(x,y)c_a + \left[1-\chi(x,y)\right]c_b(x,y)
\end{equation}
and
\begin{equation}
\label{pimage}
p(x,y)=\chi(x,y)p_a + \left[1-\chi(x,y)\right]p_b(x,y).
\end{equation}
Here $c_a$ and $p_a$ are Compton scatter and photoelectric absorption coefficients of the anomaly (object of interest) and assumed to be constant; whereas $c_b(x,y)$ and $p_b(x,y)$ are low order basis expansion representation of the background images given as
\begin{equation}
\label{cb}
c_b(x,y) = \sum_{i=1}^{N_b}\beta_i B_i(x,y)
\end{equation}
and
\begin{equation}
\label{pb}
p_b(x,y) = \sum_{i=1}^{N_b}\alpha_i B_i(x,y)
\end{equation}
where $\{\beta_i\}$ and $\{\alpha_i\}$, $i=1,\hdots,N_b$ are the sets of Compton and photoelectric expansion coefficients to be determined and $\{B_i(x,y)\}$ is the set of predefined basis functions. The order of the expansion, $N_b$, is determined depending on the desired resolution. Pixels or smoother alternatives as radial basis functions can be chosen depending on the desired resolution. This approach provides us with the possibility to reduce the number of unknowns significantly and is well suited for applications where a coarser representation of the background is acceptable. Low order basis expansion formula proved to be successful for monoenergetic x-ray tomography \cite{feng2003curve,r1992alternatives}.

The characteristic function $\chi(x,y)$ is defined to be the zero level set of an Lipschitz continuous object function $\mathcal{O}:\mathcal{D}\longrightarrow \mathbb{R}$ such that $\mathcal{O}(x,y)>0$ in
$\Omega$,  $\mathcal{O}(x,y)<0$ in  $\Omega \backslash \mathcal{D}$ and $\mathcal{O}(x,y)=0$ in
$\partial \Omega$.
%
%
Using $\mathcal{O}(x,y)$, $\chi(x,y)$ is written as
\begin{equation}
\label{chi1}
\chi(x,y)=H(\mathcal{O}(x,y))
\end{equation}
where $H$ is the step function. In practice, $H_{\epsilon}$ and its derivative $\delta_{\epsilon}$ which are smooth approximations of the step function and Dirac delta function respectively, are used \cite{osher&fedkiw}. For $\epsilon \in \mathbb{R}^+ $ we have the following:
\begin{equation}
\label{step}
H_{\epsilon}(x) = \left \{ \begin{array}{cl}
\frac{1}{2}\left(1+\frac{2}{\epsilon}+\frac{1}{\pi} sin\left(\frac{\pi x}{\epsilon}\right)\right) & \mbox{if $\vert x \vert \leq \epsilon$ } \\
1 & \mbox{if $x>\epsilon$ } \\
0 &  \mbox{if $x<-\epsilon$}
\end{array} \right.
\end{equation}
and
\begin{equation}
\label{dirac}
\delta_{\epsilon}(x) = \left \{ \begin{array}{cl}
\frac{1}{2\epsilon}\left(1+sin\left(\frac{\pi x}{\epsilon}\right)\right) & \mbox{if $\vert x \vert \leq \epsilon$} \\
0 &  \mbox{otherwise}.
 \end{array} \right.
\end{equation}
This kind of implicit (i.e. level set) representation of regions is a well-known approach successfully applied to image processing \cite{chan} as well as image formation problems \cite{dorn2000shape}. The object function $\mathcal{O}(x,y)$, is represented parametrically using a predefined basis set as
\begin{equation}
\label{O2}
\mathcal{O}(x,y) = \sum_{i=1}^L a_ip_i(x,y)
\end{equation}
%
%
where $a_i$'s are the weight coefficients whereas $p_i(x,y)$ are the functions which belong to the basis set of $\mathcal{P}=\{p_1,p_2,\dots,p_L\}$. The shape estimation problem then is reduced to the determination of a set of expansion coefficients using a quasi-Newton method (see Section 5). Consequently, we eliminate the requirement of a reinitialization process as well as implementation of narrow band methods which are essential for the standard approach where a signed distance function is used for the level set function and a curve evolution is performed. Additionally, well known properties of the level set approach of such as topological flexibility and capability to represent multiple objects is maintained \cite{osher&fedkiw}.
%

In this work, we used exponential radial basis functions (RBF's) for the background basis set $\{B_i(x,y) \vert i=1,\hdots,N_b \}$ and object function basis set $\{p_i(x,y) \vert i=1,\hdots,L \}$. Exponential RBFs are defined as
\begin{equation}
\label{RBFs}
\Phi_i(\mathbf{r}) = \exp\left(-\frac{\Vert\mathbf{r}-\mathbf{r}_i\Vert^2}{\sigma^2}\right).
\end{equation}
Here $\Phi_i:\mathcal{R}^2\rightarrow\mathcal{R}$ is the i$^{\mbox{th}}$ RBF,  $\mathbf{r}=(x,y)$ is the Cartesian coordinates vector, $\mathbf{r_i}$ is the center of i$^{\mbox{th}}$ basis function and $\sigma$ is the width which is fixed for all RBFs in the basis set. Once the number of elements in the basis set is determined $\sigma$'s can be determined by a least squares fit to simple geometries \cite{schweiger2003image}.

Let us gather all the parameters of our model in vector $\boldsymbol{\theta}^T=\left[c_a,p_a,\mathbf{a}^T,\boldsymbol{\beta}^T,\boldsymbol{\alpha}^T\right]$ where $\boldsymbol{a} = [a_1,\ldots,a_L]^T$, $\boldsymbol{\alpha} = [\alpha_1,\ldots,\alpha_{N_b}]^T$,\ and  $\boldsymbol{\beta} = [\beta_1,\ldots,\beta_{N_b}]^T$. Now, for the set of model parameters $\boldsymbol{\theta}$, the high and low energy projection models given in (\ref{projectiondL}) and (\ref{projectiondH}) can be expressed in a operator form: $K(\boldsymbol{\theta}) =\left[K(\boldsymbol{\theta})_L^T,K(\boldsymbol{\theta})_H^T\right]^T$. We define the matrix $\mathbf{B}\in \mathcal{R}^{N_p\times N_b}$ as the discretized basis matrix where $ [\mathbf{B}]_{ij} $ is the value of $B_j(x,y)$ at the center of $i^{\mbox{th}}$ pixel. Now, the vector of Compton background and photoelectric background images can be expressed as $\mathbf{B}\boldsymbol{\beta}$ and  $\mathbf{B}\boldsymbol{\alpha}$ respectively. Finally, we denote the lexicographical order of discretized object function $\mathcal{O}(x,y)$ with the vector $\boldsymbol{\mathcal{O}}\in \mathcal{R}^{N_p}$.
%
%
%
%

\section{Reconstruction Algorithm}
\label{sec:reconstruction_algorithm}

\begin{figure}
\centering
\includegraphics[width=5.5 cm]{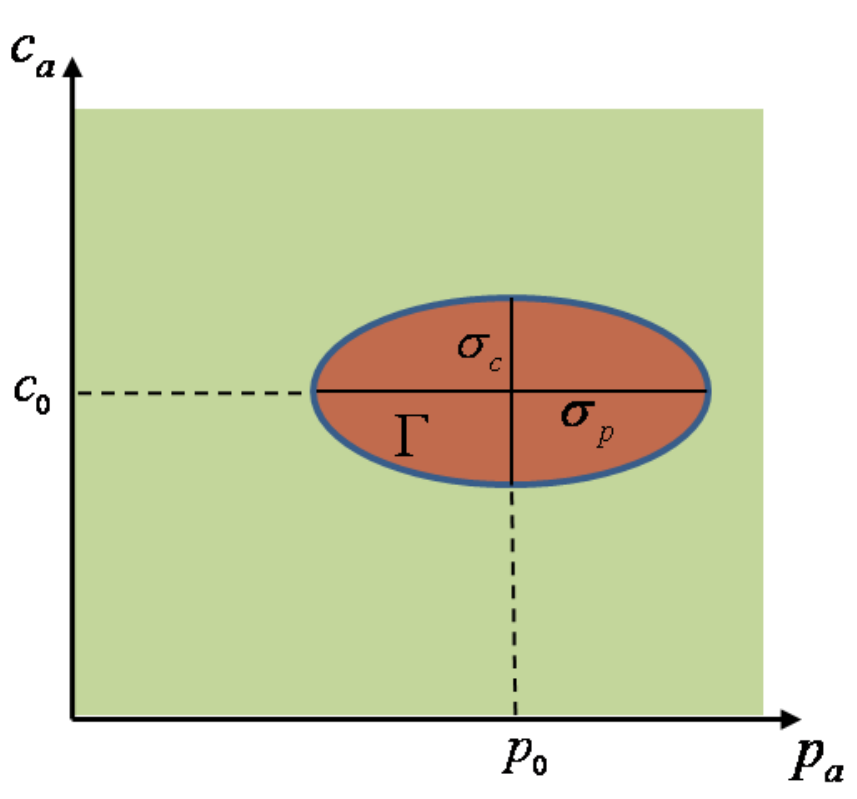}
\caption{Region $\Gamma$ for object of interest parameters }
\label{fig:cp_plane}
\end{figure}

In our variational framework, we seek for  an estimate $\hat {\boldsymbol{\theta}}$ of the unknown parameter vector $\boldsymbol{\theta}$ minimizing an objective functional $F(\boldsymbol{\theta})$ which is composed of a data misfit term and regularization terms which incorporate prior information, subject to a set of non-linear constraints as

\begin{equation}
\label{optF1}
\begin{array}{ll}
\mbox{minimize} & \begin{split} F(\boldsymbol{\theta},\boldsymbol{\lambda})& =  \frac{1}{2}\bigl(K(\boldsymbol{\theta})-\mathbf{m}\bigr)^T\Sigm\bigl(K(\boldsymbol{\theta})-\mathbf{m}\bigr)\\ &\quad + R(\boldsymbol{\theta},\boldsymbol{\lambda})
\end{split} \\
\mbox{subject to} & (p_a,c_a) \in \Gamma \\
& ([\mathbf{B}\boldsymbol{\beta}]_i,[\mathbf{B}\boldsymbol{\alpha}]_i)\notin \Gamma, \quad \mbox{for} \quad i=1,\hdots,N_p.\\
\end{array}
\end{equation}
Here the constraints account for our prior knowledge about the physical properties of the objects. We interpret this type of prior information as providing a subset $\Gamma$ of allowable values for the contrast of the object in the $c_a-p_a$ parameter space which also corresponds to pairs of values that cannot be found in the reconstruction of the background. While in general the set of allowable parameter values could be comprised of a number of disconnected regions (as would be the case where there were a number of possible chemical compounds of interest each characterized by its own uncertainty set), for simplicity here we consider the case where the constraint set is elliptical in shape. We define $\Gamma\in\mathcal{R}^2$ as an ellipse with the center $(c_0,p_0)$ major axis $\sigma_p$ and minor axis $\sigma_c$ (see Fig.
\ref{fig:cp_plane} for a symbolic representation).
Now, we can write two sets of non-linear inequality constraints $g_1(p_a,c_a):\mathcal{R}^2\rightarrow\mathcal{R}$ and $\boldsymbol{g}_2(\boldsymbol{\beta},\boldsymbol{\alpha}):\mathcal{R}^{2N_b}\rightarrow\mathcal{R}^{N_p}$ regarding the object and background contrast values respectively as
\begin{equation}
\label{g1}
g_1(p_a,c_a) = \frac{(c_a-c_0)^2}{\sigma_c^2}+\frac{(p_a-p_0)^2}{\sigma_p^2} -1
\end{equation}
and
\begin{equation}
\label{g2}
\begin{split}
\boldsymbol{g}_{2,i}(\boldsymbol{\beta},\boldsymbol{\alpha}) & = -\frac{([\mathbf{B}\boldsymbol{\beta}]_i-c_0)^2}{\sigma_c^2}-\frac{([\mathbf{B}\boldsymbol{\alpha}]_i-p_0)^2}{\sigma_p^2} +1 \\ & \quad,\mbox{for} \quad i=1,\hdots,N_p.
\end{split}
\end{equation}
The optimization problem with these inequality constraints can be rewritten as
\begin{equation}
\label{optF2}
\begin{array}{ll}
\mbox{minimize} & \begin{split} F(\boldsymbol{\theta},\boldsymbol{\lambda}) &=  \frac{1}{2}\bigl(K(\boldsymbol{\theta})-\mathbf{m}\bigr)^T\Sigm\bigl(K(\boldsymbol{\theta})-\mathbf{m}\bigr)\\
& \quad + R(\boldsymbol{\theta},\boldsymbol{\lambda}) \end{split} \\
\mbox{subject to} & g_1(p_a,c_a) \leq 0, \\
& \boldsymbol{g}_2(\boldsymbol{\beta},\boldsymbol{\alpha}) \leq 0.\\
\end{array}
\end{equation}
%

X-ray interactions, hence the photon count at the detector, is governed by a Poisson process \cite{beutel2000handbook}. Statistical inversion algorithms, where the exact log-likelihood of Poission distribution is used, have been successfully applied to medical imaging problems \cite{fessler&sukovic, de2001iterative, elbakri2003segmentation}. Sauer and Bouman \cite{bouman1996unified} derived the weighted least squares data misfit term given in (\ref{optF1}) and (\ref{optF2}) as a quadratic approximation of the Poisson log likelihood function which is successfully applied to dual energy CT problem \cite{sukovic}. In their approach the error terms are weighted by the number of counts at the detector such that $\Sigm = \mbox{diag}(\mathbf{m})$.  In transmission tomography measurements with low intensities (low photon counts) correspond to rays coincide with highly attenuating objects and have low SNR.  Therefore, the measurements with high intensities are considered to be more reliable and highly weighted. From a different point of view, if the photon count is large as will be the case here, the Poisson distribution can be approximated by a Gaussian distribution \cite{tang2009performance, siltanen2003statistical}. Furthermore, energy integrating detectors used in commercial CT scanners cause the statistics to deviate from Poisson distribution \cite{whiting2006properties} and their noise characteristics are well approximated by a Gaussian distribution \cite{sukovic, massoumzadeh2005noise}. In that case one can assume that the total measurement noise is additive Gaussian with zero mean and invertible covariance matrix $\Gamma_{\mbox{noise}}$ and set $\Sigm$ equal to $\Gamma_{\mbox{noise}}^{-1}$. In this work we assume that each measurement is independent and of equal quality; hence set $\Sigm = \sigma\bf{I}$. This approach had been successfully applied to limited data CT \cite{kolehmainen2003statistical} application and we feel it is reasonable for this work since artifacts which are caused by high density materials are complications that we prefer to consider at a later time (See Section \ref{sec:conclusions}).

The regularizer $R(\boldsymbol{\theta})$ has two terms which penalizes over-sized objects of interest and allows for the use of photoelectric absorption-Compton Scatter basis model.
\begin{equation}
\label{R}
R(\boldsymbol{\theta},\boldsymbol{\lambda}) = R_1(\boldsymbol{a}) + R_2(\boldsymbol{\alpha},\boldsymbol{\beta} ).
\end{equation}

First is the commonly used penalty term \cite{chan,lesselier} that encourages objects of interest to have a small area:
\begin{equation}
\label{R1}
R_1(\boldsymbol{a}) = \lambda_1\Vert H(\boldsymbol{\mathcal{O}}) \Vert_1.
\end{equation}

\begin{figure}
\centering
\includegraphics[width=9 cm]{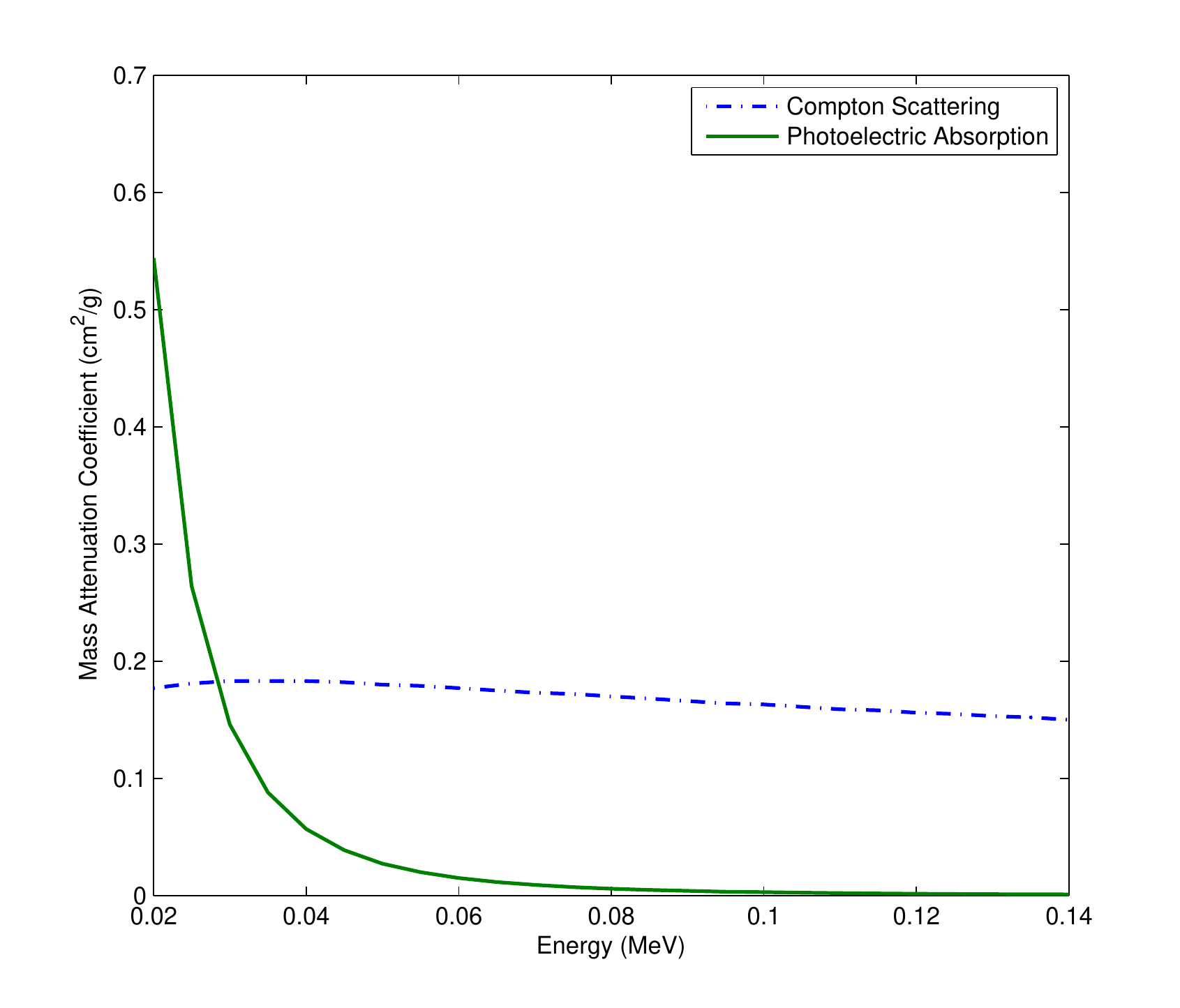}
\caption{Comparison of mass attenuation coefficients due to Compton scatter and photo electric effect for water.}
\label{fig:water_attenuation}
\end{figure}

As stated in (\ref{attenuation}) and (\ref{fp}), due to the $E^{-3}$ energy dependency of photoelectric absorption, especially for photon energies greater than $50$ KeV, the contribution of the photoelectric absorption to the total attenuation is very small compared to that of Compton scatter \cite{joseph1982effects}. Fig. \ref{fig:water_attenuation} demonstrates a comparison of the contributions of both phenomenon to the total x-ray attenuation for water as a function of energy. The cross section data for this figure is taken from Xcom: Photon cross section database \cite{berger2005xcom}. For instance at 80 KeV Compton scatter cross section is $1.7\times 10^{-1}\mbox{cm}^2/\mbox{g}$ whereas photoelectric absorption cross section is $5.77\times 10^{-3}\mbox{cm}^2/\mbox{g}$ (i.e. roughly 30 times smaller). Hence, there is severe mismatch between sensitivities of the measurement model to these components. Therefore, for noisy data, reconstruction of the photoelectric component requires additional effort especially as the total attenuation of the scene increases. In the Appendix, we perform a sensitivity analysis of linearized version of the background reconstruction problem and show that the lower bound of the error in the estimation of photoelectric absorption coefficient is significantly larger than that of the Compton scatter coefficient. One way to ameliorate such a situation is to introduce an appropriate regularization scheme.
%
%
As we are reconstructing two different parameters (Compton scatter and photoelectric absorption coefficients) of the same object, one would expect spatial co-variation between reconstructed images of these parameters. That is to say orientation of the boundaries should be similar although intensity variations within individual images may differ. We exploit this \textit{a priori} information by enforcing similarity between gradient vectors of the images of these parameters. This kind of approach was previously used in the context of geophysical imaging where Gallardo \textit{et al.} performed a joint inversion of DC resistivity and seismic refraction to characterize 2D subsurface profiles \cite{gallardo2003characterization,gallardo2005quadratic}. Their approach to enforcing structural similarity via a constraint in the optimization process required that the cross product of the gradient vectors associated with the two properties should vanish.

In our case, however, we seek structural co-variation (i.e., two images should be similar in some sense but might not be identical) between reconstructed images. Our aim is to incorporate this \textit{a priori} information to regularize our problem in a variational sense rather than imposing a constraint on the optimization process. As the boundary of the object of interest is identical in both images (i.e., $\chi(x,y)$) it suffices to deal only with background images. Herein we introduce the following correlation type of metric
\begin{equation}
\label{R2}
R_2(\boldsymbol{\alpha},\boldsymbol{\beta} ) = \lambda_2\left[\frac{\Vert \textbf{D}\textbf{B}\boldsymbol{\beta} \Vert^2_2 \Vert \textbf{D}\textbf{B}\boldsymbol{\alpha} \Vert^2_2}{\left[(\textbf{D}\textbf{B}\boldsymbol{\beta})^T(\textbf{D}\textbf{B}\boldsymbol{\alpha})\right]^2}-1\right]^2
\end{equation}
where $\textbf{D}$ is the gradient matrix defined as $\textbf{D}=[\textbf{D}_x\:\: \textbf{D}_y]$ with the first difference matrices $\textbf{D}_x$ and $\textbf{D}_y$ in $x$ and $y$ direction respectively. It is easily seen that $F_2$ decreases as the correlation of the gradients increase in negative or positive direction and vanishes when they are perfectly correlated or anti-correlated.

The constants $\left\{\lambda_i\right\} _{i=1,2}$ associated with each term in $R(\boldsymbol(\theta)$ tunes the trade off between the fidelity of the solution to the observed data, incorporation of \textit{a priori} information and regularization. Optimal choice of multiple regularization parameters is a challenging problem. L-curve \cite{belge2002efficient}, generalized cross validation(GCV) \cite{golub1979generalized} and discrepancy principle \cite{morozov1984methods} are the most frequently used methods for optimal choice of multiple regularization parameters. Although systematic selection of parameters is provided with these methods, the problem is not straightforward and requires running the optimization code several times. In this work, we tuned the regularization parameters $\lambda_1$ and $\lambda_2$ by hand such that the values that give the best results judged by the eye were assigned. Empirically speaking, for values bigger then an optimum value of $\lambda_2$, reconstruction quality remains the same while number iterations to obtain that quality increases, therefore the process becomes more time consuming.

The solution to (\ref{optF2}) is obtained using an exact penalty approach where the constrained problem is transformed to an unconstrained one so that the transformed cost functional has the same minima as the original one. This is achieved by adding a term penalizing infeasible regions to the objective functional,  then performing minimization  using an appropriate unconstrained optimization technique \cite{luenberger2008linear}. Specifically, in this work we used the following penalty function
\begin{equation}\label{penalty}
P(r,c_a,p_a,\boldsymbol{\beta},\boldsymbol{\alpha}) = r\left(g^+_1(c_a,p_a) +  \sum_{i=1}^{N_p}\boldsymbol{g}_{2,i}^+(\boldsymbol{\beta},\boldsymbol{\alpha}) \right)
\end{equation}
where $f^+(x) =\mbox{max}(f(x),0)$ and $r$ is the penalty parameter. As differentiability of the penalty term will be required we use the following smooth approximation \cite{yamashita2007primal} of max$(x,0)$ function:
\begin{equation}\label{smax}
\mbox{max}(x,0) \approx \frac{1}{2}(\sqrt{x^2+\epsilon}+x)
\end{equation}
where $\epsilon\in \mathcal{R^+}$ is small.

According to exact penalty theorem  \cite{luenberger2008linear} If $\boldsymbol{\mu}$ are the Lagrange multipliers of the original problem, for $r>\mbox{max}(\mu_i:i=1,\dots,1+N_p)$ the local minima of (\ref{optF2}) is also the local minima of the following unconstrained problem:
\begin{equation}
\label{Faug}
\begin{split}
F_p(\boldsymbol{\theta},\boldsymbol{\lambda}, r) & =  \frac{1}{2}\bigl(K(\boldsymbol{\theta})-\mathbf{m}\bigr)^T\Sigm\bigl(K(\boldsymbol{\theta})-\mathbf{m}\bigr)\\
& \quad + R(\boldsymbol{\theta}) + P(r,c_a,p_a,\boldsymbol{\beta},\boldsymbol{\alpha}).
\end{split}
\end{equation}
As opposed to other methods to solve constraint optimization problems such as quadratic penalty method where one needs to solve sequence of sub-problems with increasing penalty parameter, it suffices to find one $r$ that will be fixed during the minimization process. Finding the Lagrange multipliers, however, is a hard problem. Nonetheless, the value of the penalty parameter $r$ can be determined relatively easier than $\lambda_i$'s as the solution is not prone to fail with changes in $r$ and it suffices to, heuristically, set it large enough so that the constraints are satisfied \cite{yamashita2007primal}. We found this approach satisfactory for the purpose and the minimization algorithm (Levenberg-Marquardt) of this work. For an adaptive penalty parameter update strategy for a S$l_1$QP \cite{fletcher1981practical} based method, we refer the reader to the work of Byrd \emph{et al.} \cite{byrd2008steering}.

The minimization of the cost function in (\ref{Faug}) is achieved by Levenberg-Marquardt algorithm \cite{marquardt1963algorithm}. For this aim the cost function is rewritten in terms of an error vector $\epsilon$ as
\begin{equation}
\label{Feps}
F_p(\mathbf{\theta}) =  \boldsymbol{\epsilon}(\boldsymbol{\theta})^T\boldsymbol{\epsilon}(\boldsymbol{\theta}).
\end{equation}
The error term has the form
$\boldsymbol{\epsilon}(\boldsymbol{\theta})^T= \left[
\boldsymbol{\epsilon}_1^T,\boldsymbol{\epsilon}_2^T, \epsilon_3,\boldsymbol{\epsilon}_4^T\right]$ where each term is associated with the corresponding term in the cost functional and given as
\begin{equation}
\label{eps1}
\boldsymbol{\epsilon}_1 =  K(\boldsymbol{\theta})-\mathbf{m},
\end{equation}
\begin{equation}
\label{eps2}
\boldsymbol{\epsilon}_2(\mathbf{a})=
\sqrt{\lambda_1 H(\boldsymbol{\mathcal{O}})},
\end{equation}
\begin{equation}
\label{eps3}
\epsilon_3(\boldsymbol{\alpha,\beta})=\sqrt{\lambda_2}\left(\frac{\Vert D\boldsymbol{\beta} \Vert^2_2 \Vert D\boldsymbol{\alpha} \Vert^2_2}{(D\boldsymbol{\beta})^T(D\boldsymbol{\alpha})}-1\right)
\end{equation}
%
%
%
%
and
\begin{equation}
\label{eps4}
\boldsymbol{\epsilon}_4=\sqrt{r}\left [ \sqrt{g_1^+(c_a,p_a)},\sqrt{g_{2,1}^+(\boldsymbol{\alpha},\boldsymbol{\beta})},\dots, \sqrt{g_{2,N^p}^+(\boldsymbol{\alpha},\boldsymbol{\beta})} \right]^T.
\end{equation}

In order to employ Levenberg-Marquardt algorithm, calculation of the Jacobian matrix $J$, whose rows are the derivatives of $\boldsymbol{\epsilon}(\boldsymbol{\theta})$ with respect to each element in the unknown parameter vector $\mathbf{\boldsymbol{\theta}}$, is required. In detail, $n$th column of $J$ is the derivative of $\boldsymbol{\epsilon}(\boldsymbol{\theta})$ with respect to $n$th element in the parameter vector $\boldsymbol{\theta}$.
\begin{equation}
\label{jacobian}
\textbf{J} = \left[ \frac{\partial \boldsymbol{\epsilon}(\boldsymbol{\theta})}
{\partial\{c_a,p_a,\mathbf{a},\boldsymbol{\beta},\boldsymbol{\alpha}\}} \right].
\end{equation}
These derivatives are easily obtained analytically. The details are omitted here, although the Appendix provides some intuition. The solution is obtained by updating $\boldsymbol{\theta}$ each iteration as the following
\begin{equation}
\label{update}
\boldsymbol{\theta}^{n+1} = \boldsymbol{\theta}^n + \textbf{h}
\end{equation}
where $h$ is the solution of the following linear system at each iteration
\begin{equation}
\label{h}
(\textbf{J}^T\textbf{J} + \mu\textbf{I})\textbf{h} = -\textbf{J}^T
\boldsymbol{\epsilon}  \;\; \mbox{with}\;\; \mu \geq 0.
\end{equation}
Here $\textbf{I}$ is the identity matrix, $\mu$ is the damping parameter affecting the size and direction of $\mathbf{h}$ and it is adaptively tuned during the iteration process \cite{madsen2004methods}.

\section{Numerical Examples}
\label{sec:numerical_examples}

To validate our method we performed reconstructions from simulated data according to the data acquisition models given in (\ref{projectiondL}) and (\ref{projectiondH}). The background signals $\textbf{r}_{L}$ and $\textbf{r}_{H}$ are simulated as white Gaussian noise. Parallel beam measurements are simulated for equally spaced 30 angles between 0 and 180 degrees each with 150 source-detector pairs. We compare our results with the method proposed in \cite{ying} which is based on estimation of two sinograms corresponding to Compton scatter and photoelectric absorption then using a FBP method to reconstruct the final images. Their work is also based on reconstructing Compton scatter and photoelectric absorption properties with a FBP approach; therefore it gives us the opportunity to compare the performance of our method with the standard method that is proposed for luggage screening purposes. A Ram-Lak filter multiplied with a Hamming window was used in the FBP inversion. We will call this method DEFBP (dual energy FBP). We also applied thresholds to the reconstructed images to reveal regions $\boldsymbol{\chi}_t$ of object of interest defined by the region $\Gamma$ (see Fig. \ref{fig:cp_plane}). Specifically if reconstructed Compton and photoelectric images are $\mathbf{I}_c$ and $\mathbf{I}_p$ respectively then $\boldsymbol{\chi}_t$ can be calculated with the Hadamard product as
\begin{equation}
\label{eq:stop}
\boldsymbol{\chi}_t = \boldsymbol{\chi}_{\mbox{c}} \circ \boldsymbol{\chi}_{\mbox{p}}
\end{equation}
where $\boldsymbol{\chi}_{\mbox{c}}$ and $\boldsymbol{\chi}_{\mbox{p}}$ are given as
\begin{equation}
\label{chi}
[\boldsymbol{\chi}_{\mbox{c(p)}}]_{ij} = \left \{ \begin{array}{cl}
1 & \mbox{  if  }
\left\vert [\boldsymbol{\chi}_{\mbox{c(p)}}]_{ij}-c(p)_0 \right\vert \leq \sigma_{c(p)} \\
0 & \mbox{  otherwise.}  \\
\end{array} \right.
\end{equation}

In the minimization process we followed a coordinate decent type algorithm where object boundary(shape) parameters ($\{a_i\}$) are updated while contrast parameters are kept fixed and then contrast parameters ($\boldsymbol{\beta}$, $\boldsymbol{\alpha}$, $c_a$, $p_a$) are updated while level set parameters are fixed. Convergence is checked for each individual cycle using the stopping condition \cite{madsen2004methods}
\begin{equation}
\label{eq:stop}
\Vert \mathbf{x}^k - \mathbf{x}^{k-1} \Vert < \epsilon(1+\Vert\mathbf{x}^{k-1}\Vert)  \quad \mbox{or} \quad k>k_{\mbox{max}}
\end{equation}
where $\mathbf{x}^k$ corresponds to vector of shape or contrast parameters at $k^{\mbox{th}}$ iteration, $k_{\mbox{max}}$ is maximum number of iterations and $\epsilon$ is a small, positive number which is taken $10^{-6}$. Once both cycles are completed the overall convergence is checked using the same criteria given in (\ref{eq:stop}) on the reconstructed photoelectric and Compton image vectors.

\begin{figure}
\centering
\includegraphics[width=3.7in]{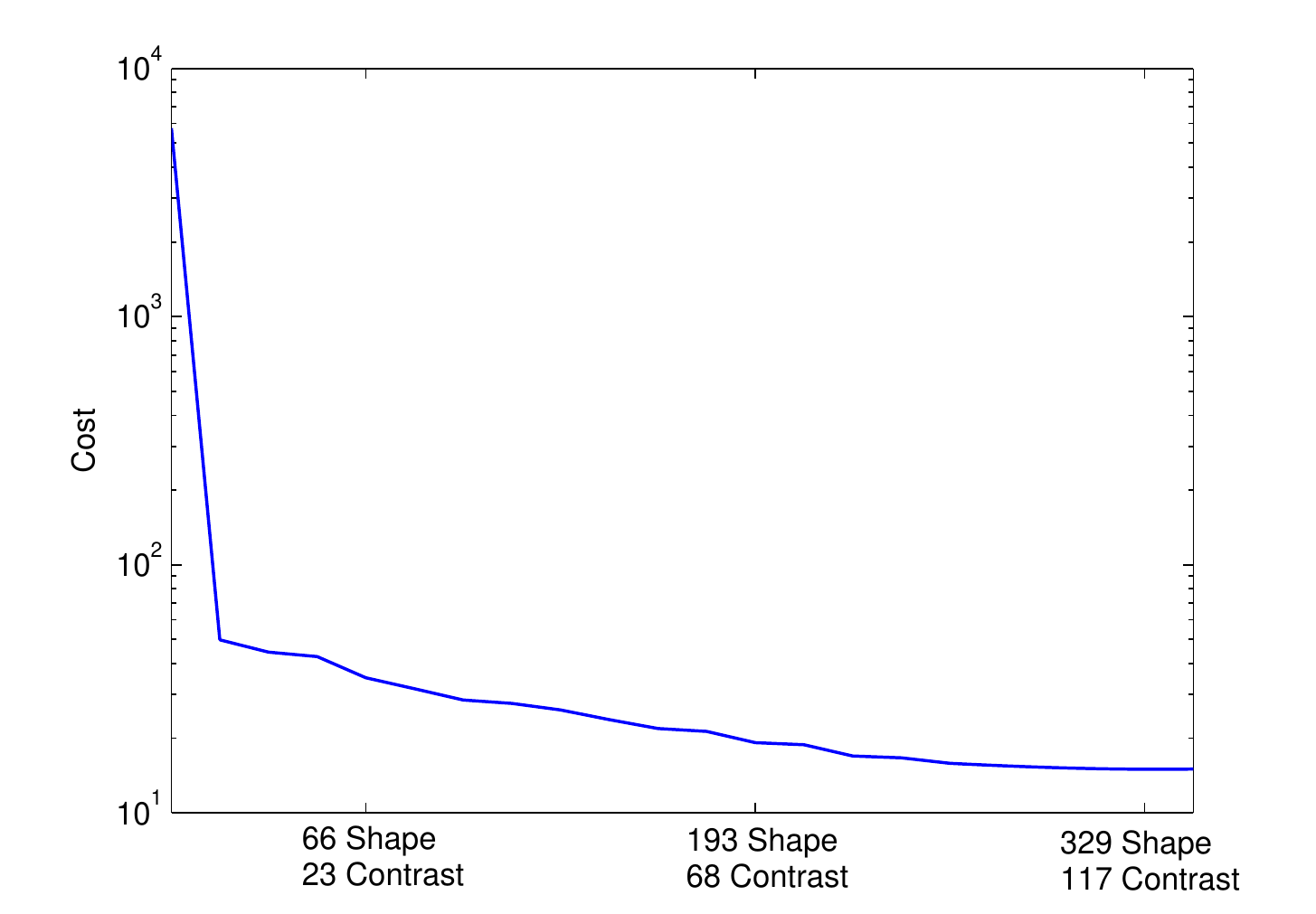}
\caption{The value of the cost function $F$, versus number of iterations for the example shown in Fig. \ref{fig:60dBCorr}.}
\label{fig:cost_evolution}
\vspace*{-8mm}
\end{figure}

Quantitative accuracy of reconstructions are determined by relative $L^2$-error which is given as
\begin{equation}
E_{L^2} = \frac{\parallel \hat{\mathbf{I}}-\mathbf{I} \parallel^2_2}{\parallel \mathbf{I} \parallel^2_2  }.
\end{equation}
Here, $\hat{\mathbf{I}}$ is the reconstruction of either the Compton image or the photoelectric image and $\mathbf{I}$ is the corresponding true image. The accuracy of the boundary reconstruction $\hat{\boldsymbol{\chi}}$ compared to the true boundary $\boldsymbol{\chi} $ is determined using the Dice's coefficient. Assuming $\hat{\boldsymbol{\chi}}$ and $\boldsymbol{\chi}$ are sets of points in space where the points are corresponding pixels of the object of interest, Dice's coefficient is given as
\begin{equation}
D_{\boldsymbol{\chi}} = \frac{2|\hat{\boldsymbol{\chi}} \cap \boldsymbol{\chi}|}{|\hat{\boldsymbol{\chi}}| + |\boldsymbol{\chi}|}.
\end{equation}
Here $D_{\boldsymbol{\chi}} = 1$ shows maximum similarity (i.e., the sets are identical) and $D_{\boldsymbol{\chi}} = 0$ shows the sets do not have a common point (i.e., the sets are disjoint).

For all examples, $N_b=676$ RBF's with $\sigma_b=6\Delta x$ where $\Delta x$ is the length of one pixel were used for background representation given in (\ref{cb}) and (\ref{pb}). This corresponds to a RBF for every 4x4 super-pixel and 676 unknown coefficients for each background image instead of 10$^4$ which would have been the case for a pixel based reconstruction. For the level set function we used $L=144$ RBF's with $\sigma_s=10\Delta x$ as the basis set. The level set function is initialized randomly whereas background parameters $\boldsymbol{\beta}$ and $\boldsymbol{\alpha}$ are set to $8\times 10^{-3}$ and $80$ respectively. These values are chosen so that the contrast of initial background images were small. The regularization parameters, $\lambda_1$ and $\lambda_2$ were set to 0.1 and 10 respectively; whereas the penalty parameter, $r$ was set to $10^5$.

\begin{figure*}[t!]
\centering
$\begin{array}{c@{\hspace{0in}}c@{\hspace{0in}}c}
\includegraphics[width=2.1in, trim = 5mm 1mm 5mm 0mm, clip=true]{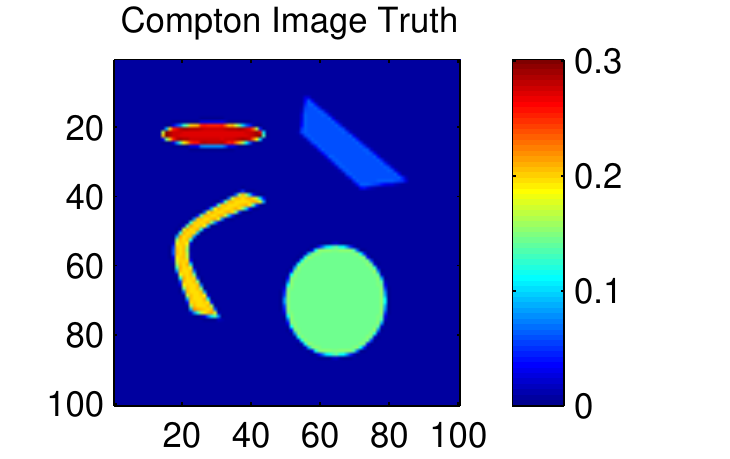}  &
\includegraphics[width=2.1in, trim = 5mm 1mm 5mm 0mm, clip=true]{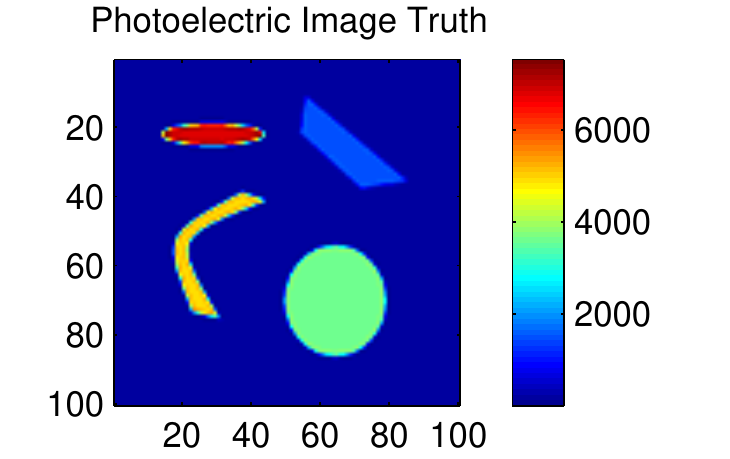} &
\includegraphics[width=1.8in, trim = 14mm 1mm 6mm 0mm, clip=true]{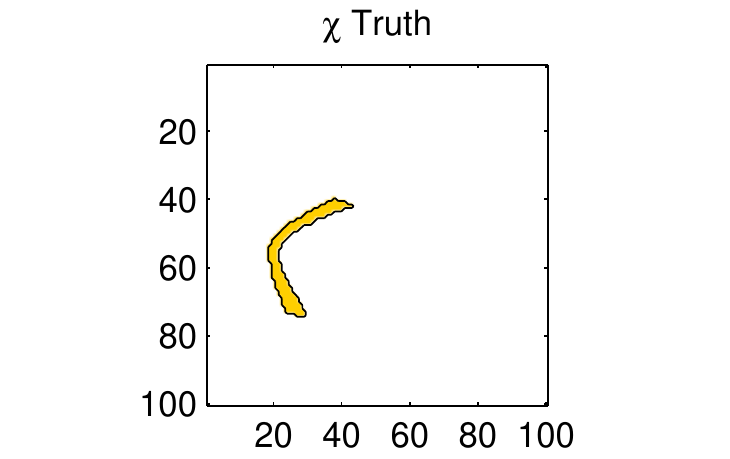}   \\

\includegraphics[width=2.1in, trim = 5mm 1mm 5mm 0mm, clip=true]{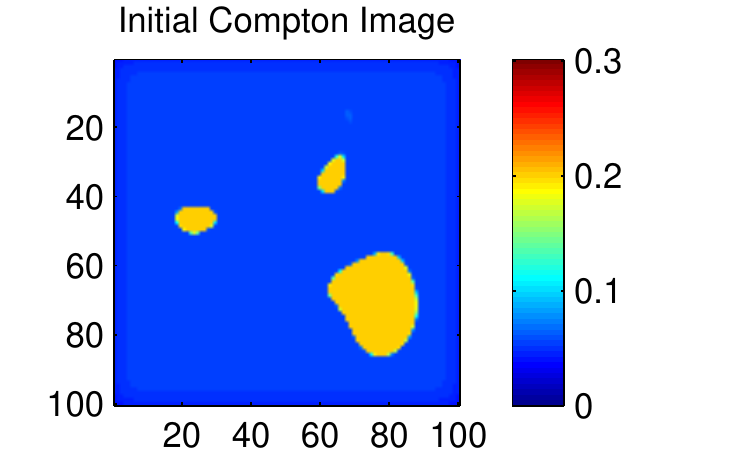}  &
\includegraphics[width=2.1in, trim = 5mm 1mm 5mm 0mm, clip=true]{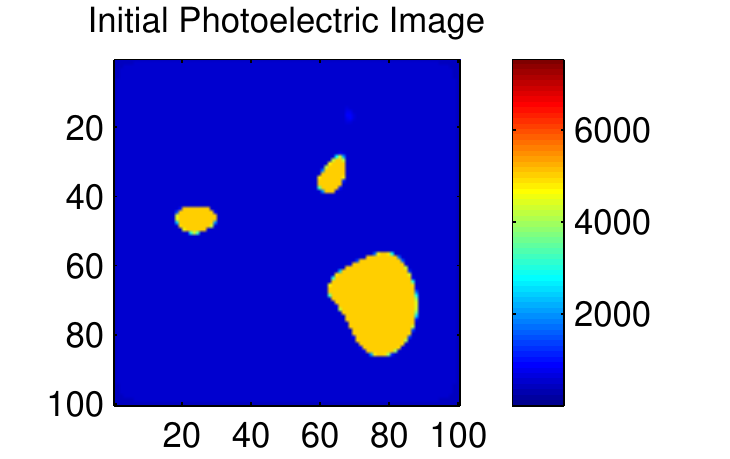} &
\includegraphics[width=1.8in, trim = 14mm 1mm 6mm 0mm, clip=true]{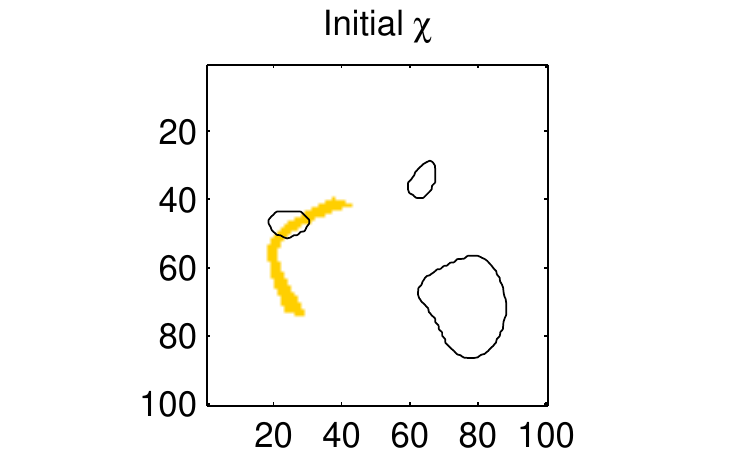}   \\

\includegraphics[width=2.1in, trim = 5mm 1mm 5mm 0mm, clip=true]{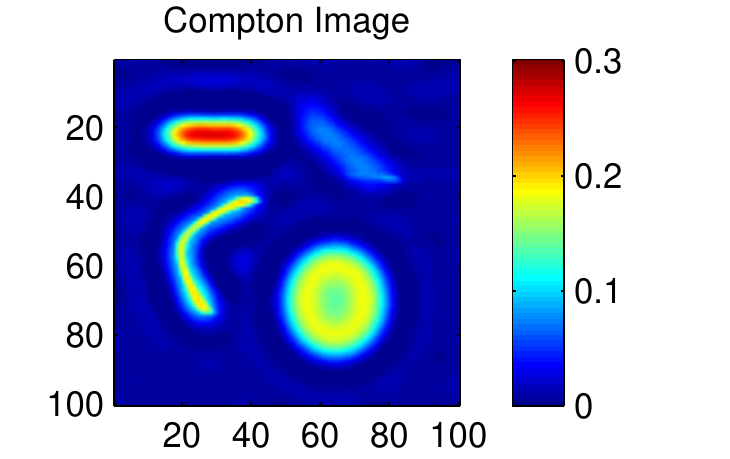}  &
\includegraphics[width=2.1in, trim = 5mm 1mm 5mm 0mm, clip=true]{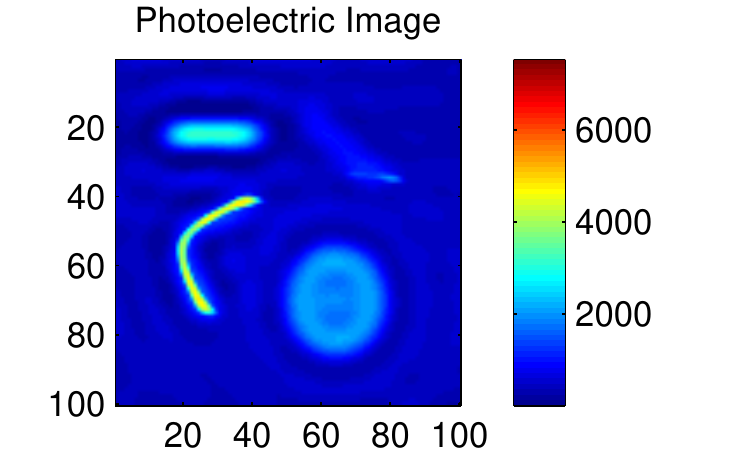} &
\includegraphics[width=1.8in, trim = 14mm 1mm 6mm 0mm, clip=true]{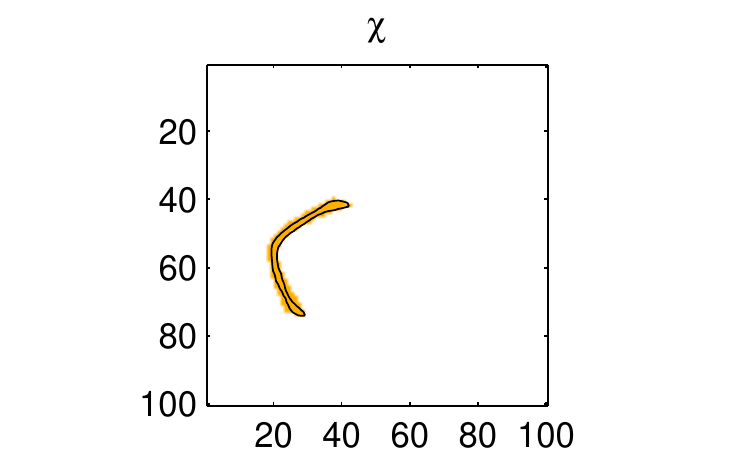}   \\

\includegraphics[width=2.1in, trim = 5mm 1mm 5mm 0mm, clip=true]{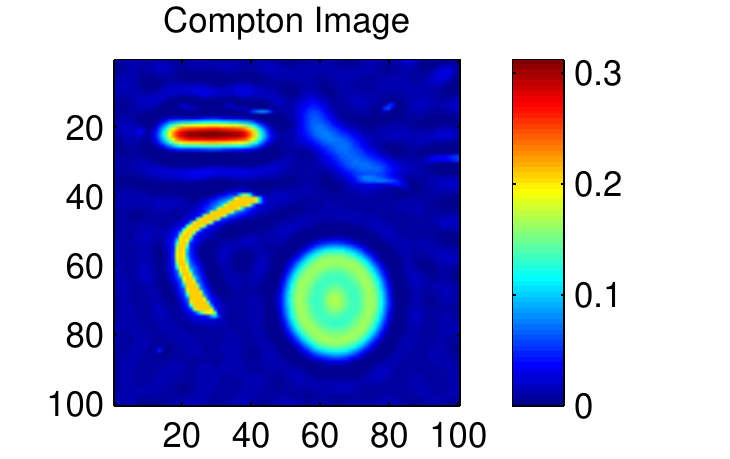}  &
\includegraphics[width=2.1in, trim = 5mm 1mm 5mm 0mm, clip=true]{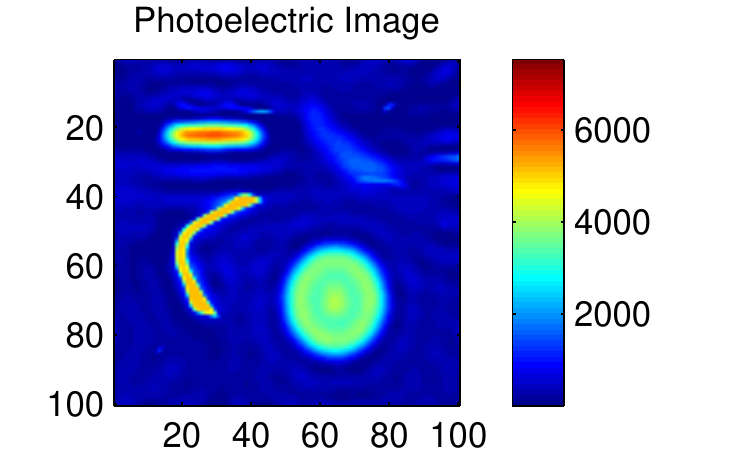} &
\includegraphics[width=1.8in, trim = 14mm 1mm 6mm 0mm, clip=true]{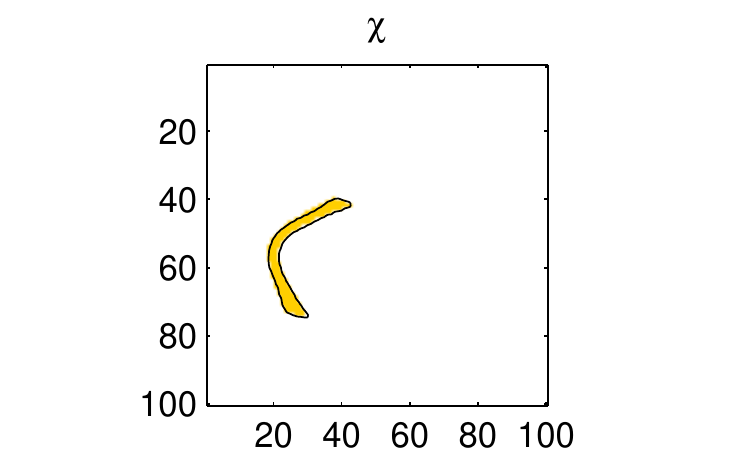}

\end{array}$
\caption{ Simulation with the first phantom with 60dB background noise. First row: Ground truth images. Second row: initial images. Third row: Reconstruction after 3 cycles of shape and contrast updates. Fourth row: Final reconstructions after 11 cycles. First column: Compton image. Second column: Photoelectric image. Third column: Characteristic function, $\chi$, of the object. The solid yellow is the true object while the thin line represents the estimated boundary.   }
\label{fig:60dBCorr}
\end{figure*}

The first set of phantoms that consist of objects with various shapes and constant intensities on a homogeneous background have the size of 20x20 cm and discretized into 100$\times$100 pixels. We emphasize that the geometry of the photoelectric absorption and Compton scatter phantoms are taken to be same although the contrasts are different as the images represent distinct physical properties of the medium. The object of interest parameters that define $\Gamma$ are set as the following: $c_0 = 0.19$, $\sigma_c = 0.05$, $p_0 = 5000$, $\sigma_p = 500$; and the boomerang shaped object on lower left is assigned as an object of interest (see first row of Fig. \ref{fig:60dBCorr}) with $c_a = 0.2$ and $p_a=5000$. These parameters are in the range that correspond to real materials that may be present in a luggage  and are chosen generically to test our algorithm on objects with moderate attenuation properties.

\begin{figure*}[t!]
\centering
$\begin{array}{c@{\hspace{.01in}}c@{\hspace{.01in}}c}

\includegraphics[width=2.1in, trim = 5mm 1mm 5mm 0mm, clip=true]{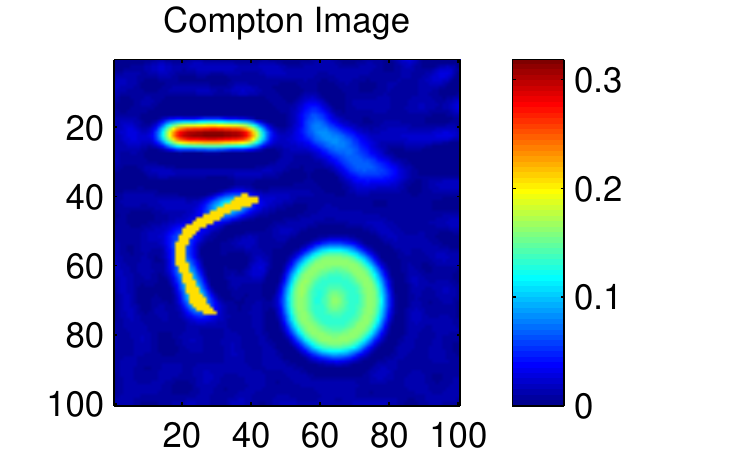}  &
\includegraphics[width=2.1in, trim = 5mm 1mm 5mm 0mm, clip=true]{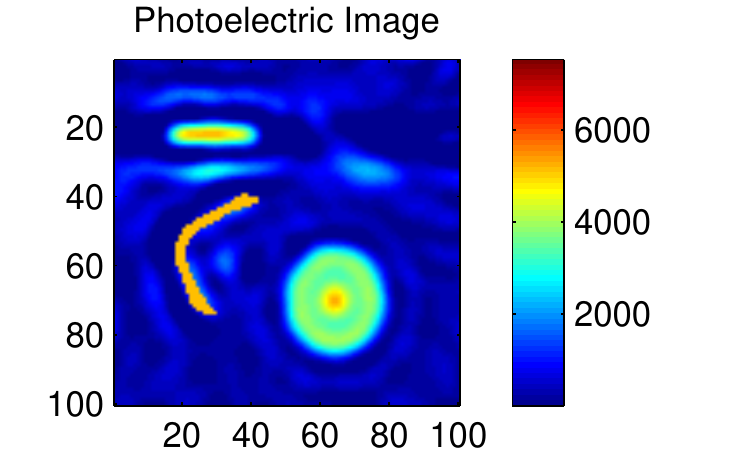} &
\includegraphics[width=1.8in, trim = 14mm 1mm 6mm 0mm, clip=true]{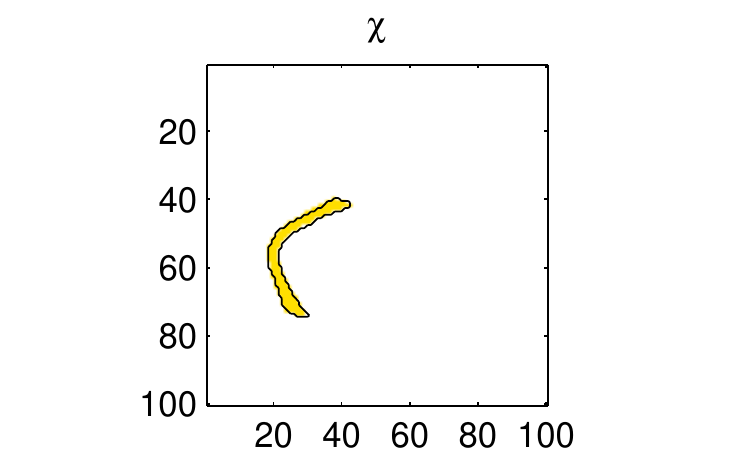}  \\

\includegraphics[width=2.1in, trim = 5mm 1mm 5mm 0mm, clip=true]{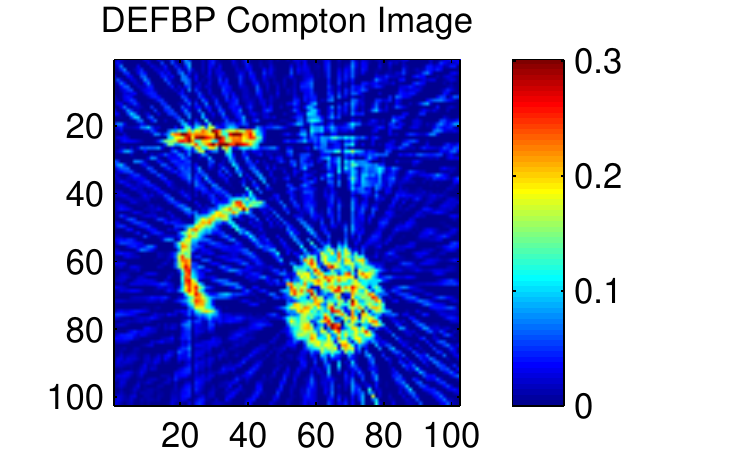}  &
\includegraphics[width=2.1in, trim = 5mm 1mm 5mm 0mm, clip=true]{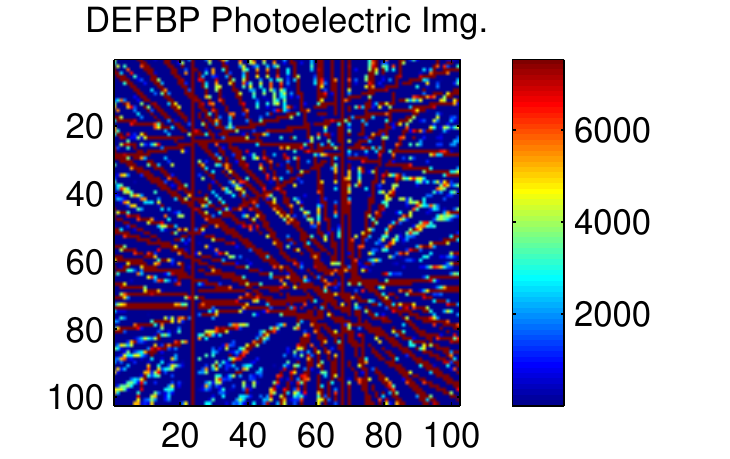} &
\includegraphics[width=1.8in, trim = 14mm 1mm 6mm 0mm, clip=true]{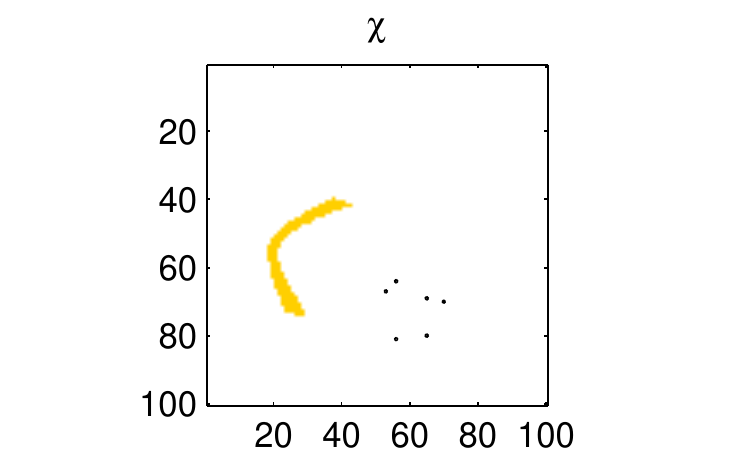}  \\

\includegraphics[width=2.1in, trim = 5mm 1mm 5mm 0mm, clip=true]{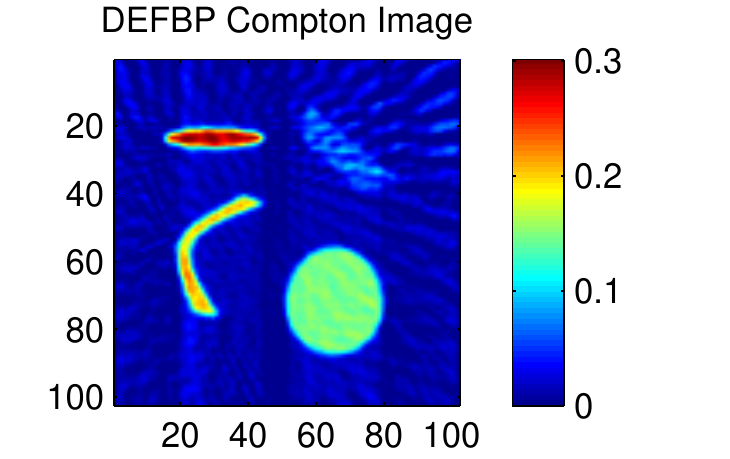}  &
\includegraphics[width=2.1in, trim = 5mm 1mm 5mm 0mm, clip=true]{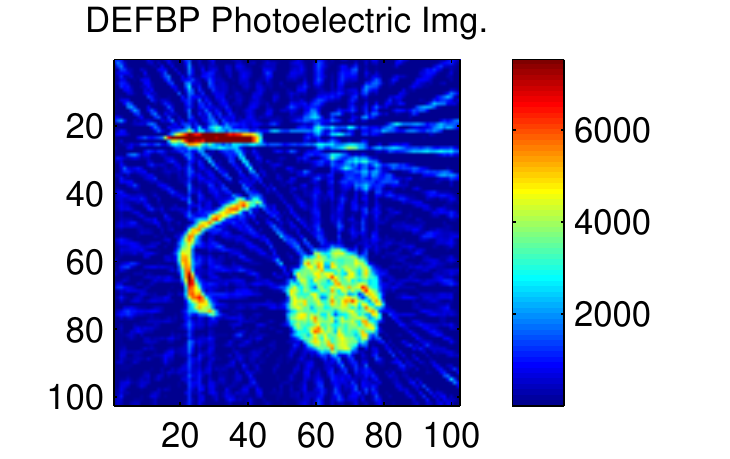} &
\includegraphics[width=1.8in, trim = 14mm 1mm 6mm 0mm, clip=true]{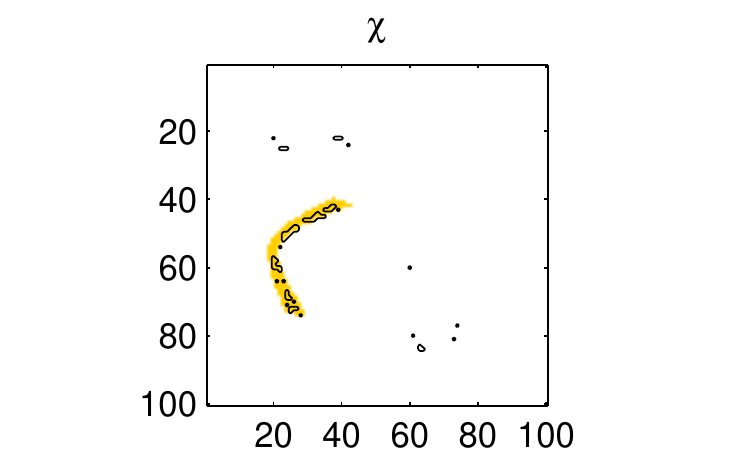}

\end{array}$
\caption{ Simulation with the first phantom with 60dB background noise. First row: Proposed method without regularizer, $R_2$. Second row: DEFPB reconstruction. Third Row: DEFBP reconstruction without background noise. First column: Compton image.  Second column: Photoelectric image. Third column: Characteristic function, $\chi$, of the object. The solid yellow is the true object while the thin line represents the estimated boundary.}
\label{fig:60dB}
\end{figure*}

In the first example the first phantom was used and dual enery measurements are simulated for 60dB background noise. Fig. \ref{fig:cost_evolution} shows the value of the cost function $F_p(\boldsymbol{\theta},\boldsymbol{\lambda}, r)$ during minimization process. Fig. \ref{fig:60dBCorr} shows the actual phantoms for Compton and photoelectric images, initializations for both images as well as the object boundary, reconstructions after 3 cycles of shape and contrast updates and the end results after 11 cycles. In the rightmost column, the object boundaries are shown with black solid lines where the yellow object shows the support of the true object of interest. Fig. \ref{fig:60dB} shows the reconstruction using the proposed method without using the correlation based regularizer $R_2$ and DEFB reconstructions with 60dB background noise and without background noise.
\begin{figure*}[t!]
\centering
$\begin{array}{c@{\hspace{.01in}}c@{\hspace{.01in}}c}

\includegraphics[width=2.1in, trim = 5mm 1mm 5mm 0mm, clip=true]{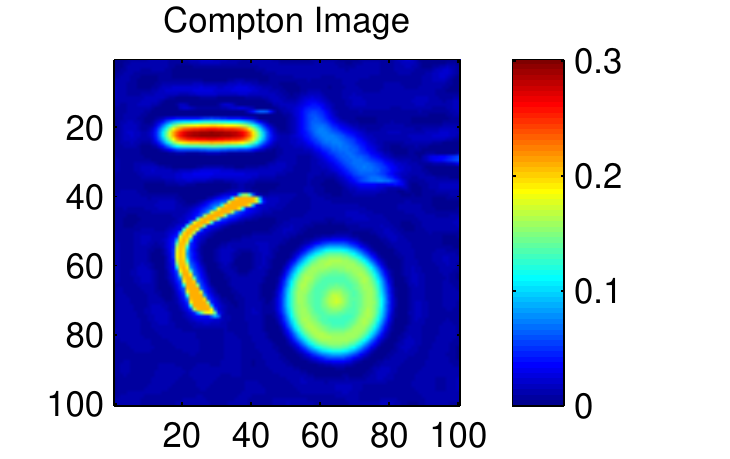}  &
\includegraphics[width=2.1in, trim = 5mm 1mm 5mm 0mm, clip=true]{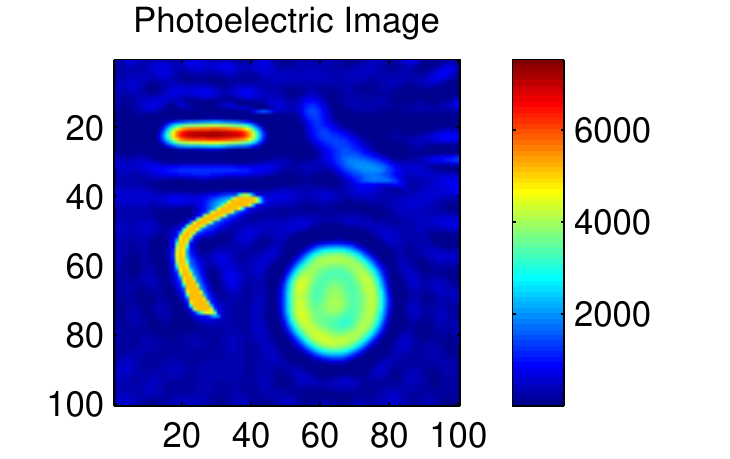} &
\includegraphics[width=1.8in, trim = 14mm 1mm 6mm 0mm, clip=true]{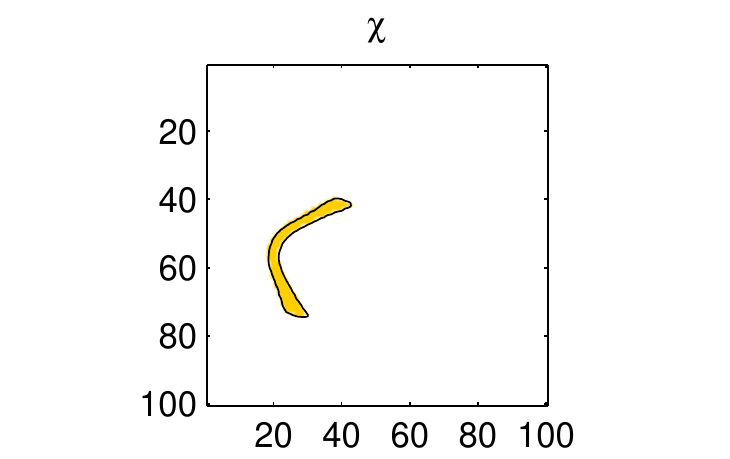} \\

\includegraphics[width=2.1in, trim = 5mm 1mm 5mm 0mm, clip=true]{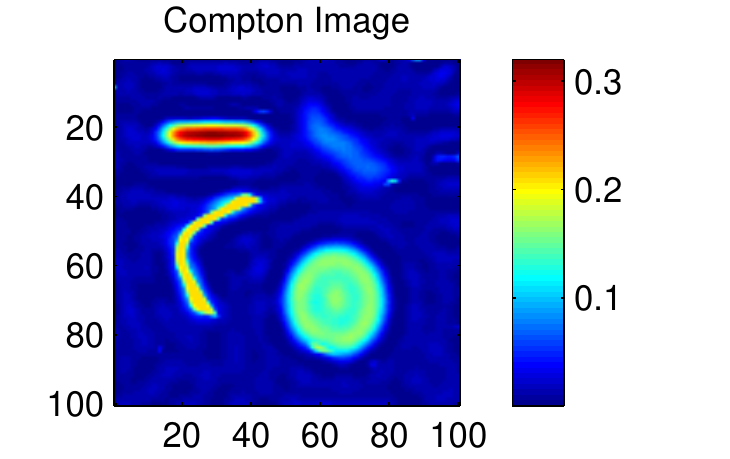}  &
\includegraphics[width=2.1in, trim = 5mm 1mm 5mm 0mm, clip=true]{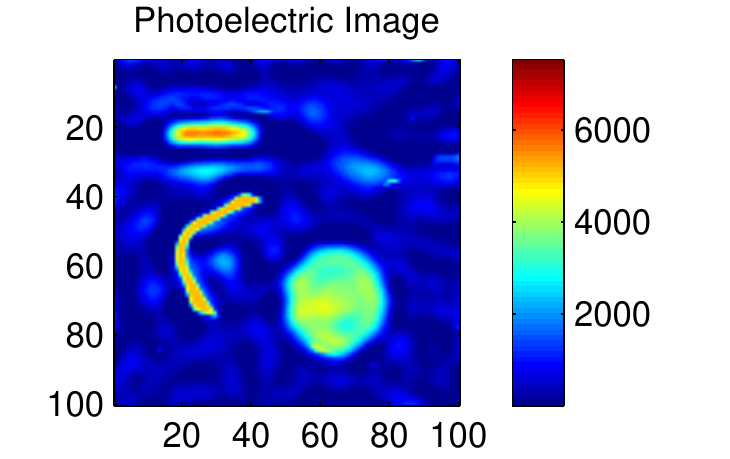} &
\includegraphics[width=1.8in, trim = 14mm 1mm 6mm 0mm, clip=true]{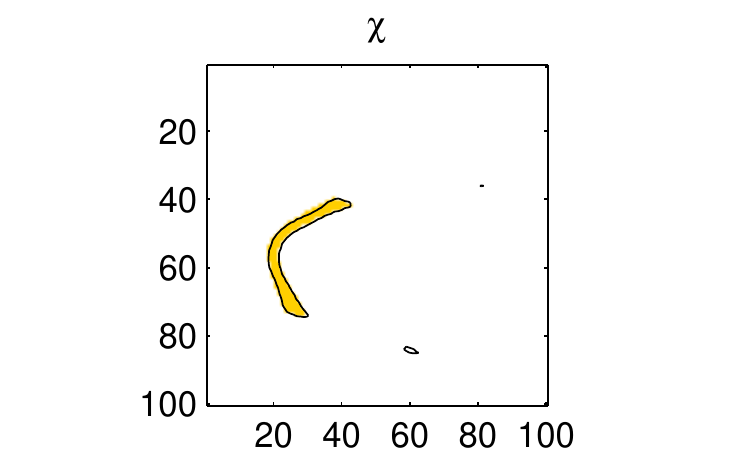} \\

\includegraphics[width=2.1in, trim = 5mm 1mm 5mm 0mm, clip=true]{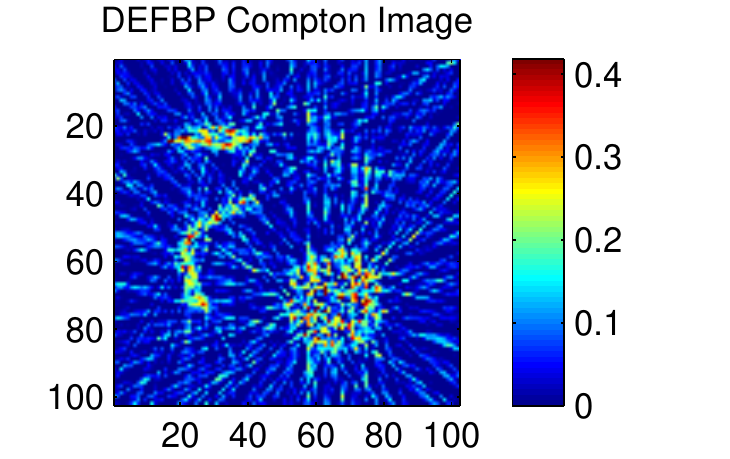}  &
\includegraphics[width=2.1in, trim = 5mm 1mm 5mm 0mm, clip=true]{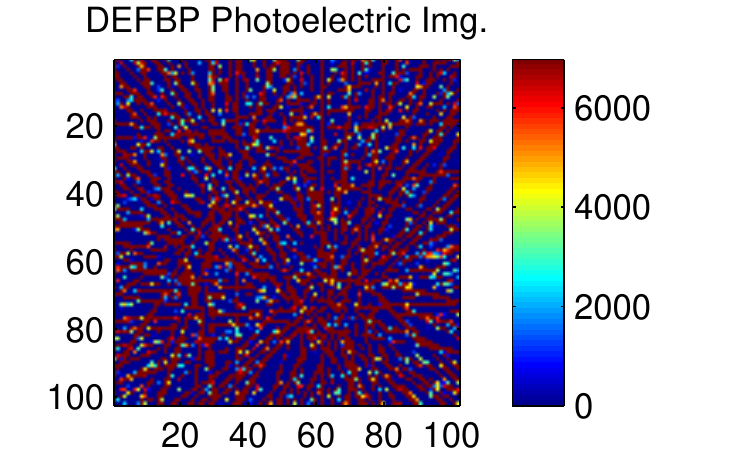} &
\includegraphics[width=1.8in, trim = 14mm 1mm 6mm 0mm, clip=true]{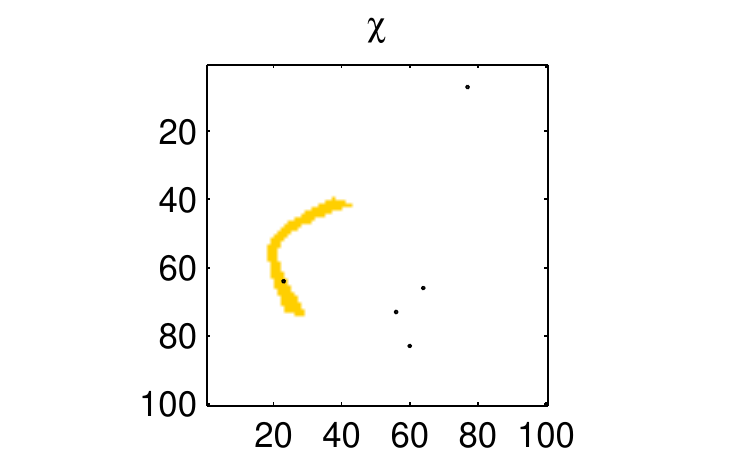}

\end{array}$

\caption{Simulation with the first phantom with 40dB background noise. First row: Proposed method. Second row: Proposed method without regularizer, $R_2$. Third row: DEFPB reconstruction. First column: Compton image. Second column: Photoelectric image. Third column: Characteristic function, $\chi$, of the object. The solid yellow is the true object while the thin line represents the estimated boundary.  }
\label{fig:40dB}
\end{figure*}
\begin{table}
\centering
\caption{Error analysis for the simulation using the first phantom with 60\lowercase{d}B background noise}
\vspace{-3mm}
\begin{tabular}{c c c c }
& $E_{L^2}$  & $E_{L^2}$  &
\\
Method &Compton&photoelectric&$D_{\boldsymbol{\chi}}$\\
\hline
\vspace{-2mm}
\\
Proposed method & 0.0741 & 0.0873 & 0.9387 \\
Proposed method without $R_2$ & 0.0818 & 0.1977 & 0.9283 \\
DEFBP & 0.4166 & 129.57 & 0 \\
DEFBP without b.ground noise.  & 0.1025 & 0.3011 & 0.4229 \\
\end{tabular}
\label{table:Error60dB}
\end{table}
\begin{figure*}[t!]
\centering
$\begin{array}{c@{\hspace{.01in}}c@{\hspace{.01in}}c}

\includegraphics[width=2.1in, trim = 5mm 1mm 5mm 0mm, clip=true]{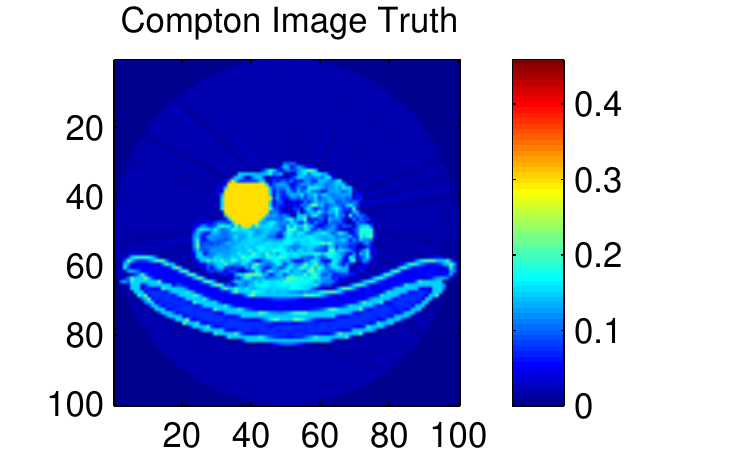}  &
\includegraphics[width=2.1in, trim = 5mm 1mm 5mm 0mm, clip=true]{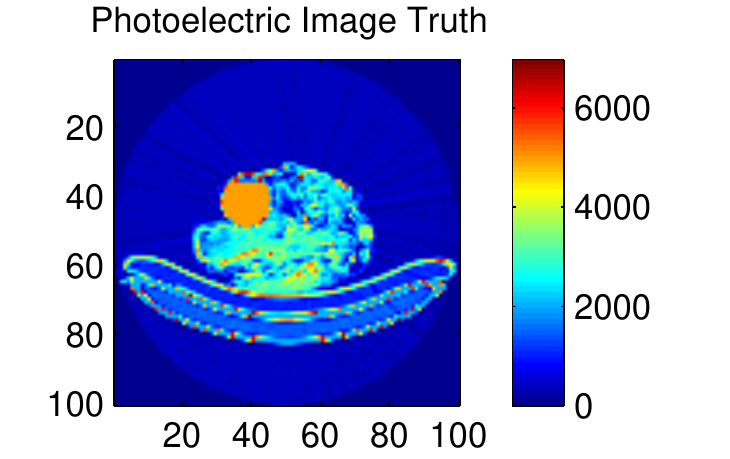} &
\includegraphics[width=1.8in, trim = 14mm 1mm 6mm 0mm, clip=true]{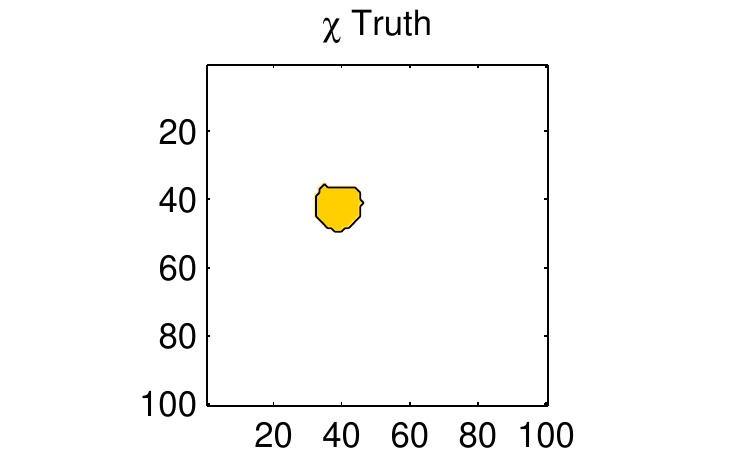}  \\

\includegraphics[width=2.1in, trim = 5mm 1mm 5mm 0mm, clip=true]{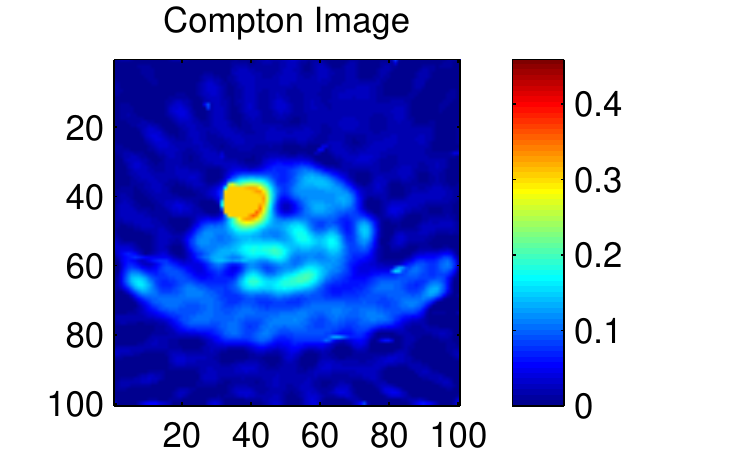}  &
\includegraphics[width=2.1in, trim = 5mm 1mm 5mm 0mm, clip=true]{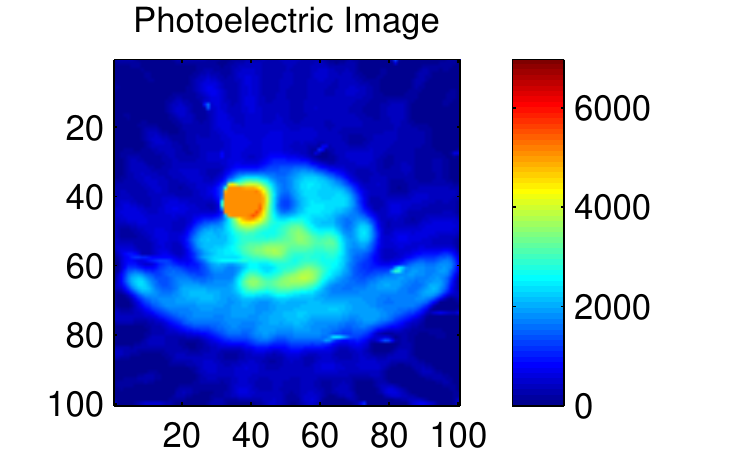} &
\includegraphics[width=1.8in, trim = 14mm 1mm 6mm 0mm, clip=true]{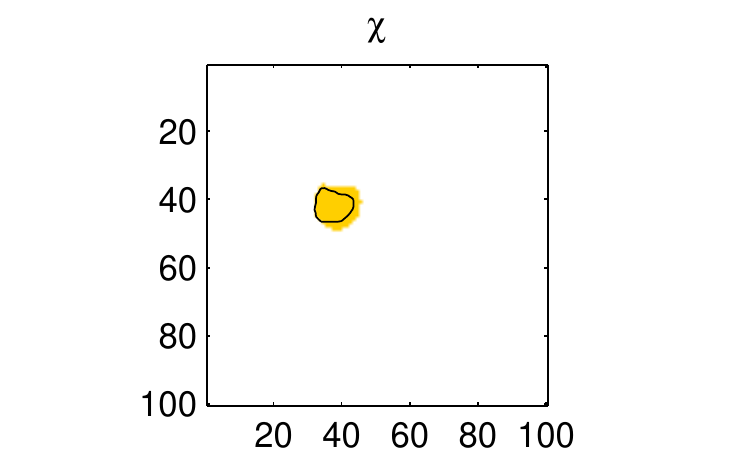}  \\

\includegraphics[width=2.1in, trim = 5mm 1mm 5mm 0mm, clip=true]{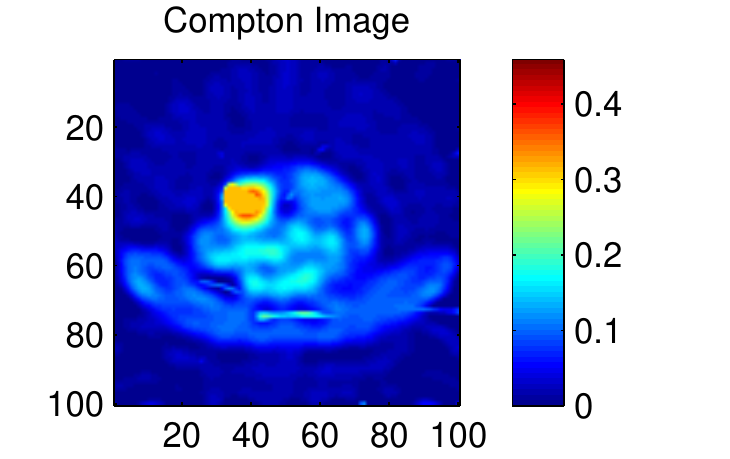}  &
\includegraphics[width=2.1in, trim = 5mm 1mm 5mm 0mm, clip=true]{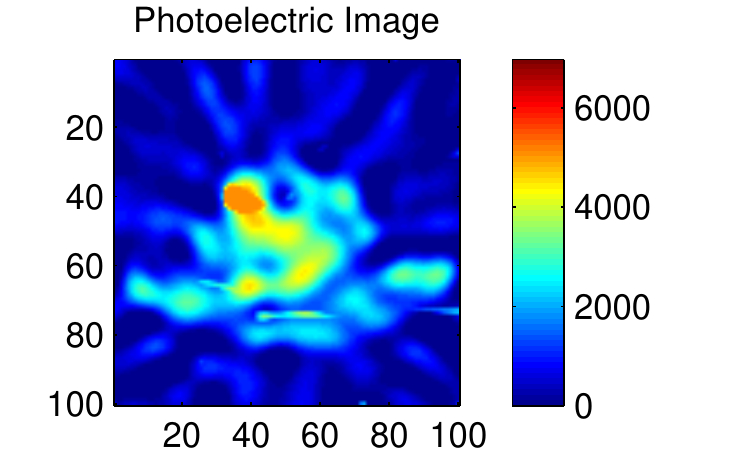} &
\includegraphics[width=1.8in, trim = 14mm 1mm 6mm 0mm, clip=true]{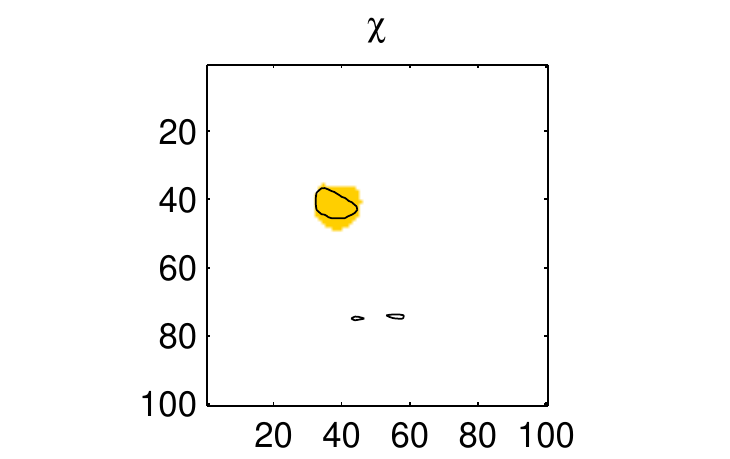}  \\

\end{array}$

\caption{ Simulation with the duffle bag phantom with 40dB background noise. First row: Ground truth images. Second row: Proposed method. Third row: Proposed method without regularizer, $R_2$. Third Row: Proposed method without regularizer, $R_2$. First column: Compton image. Second column: Photoelectric image. Third column: Characteristic function, $\chi$, of the object. The solid yellow is the true object while the thin line represents the estimated boundary.}
\label{fig:Duffle}
\end{figure*}
\begin{table}
\centering
\caption{Error analysis for the simulation using the first phantom with 40dB background noise}
\vspace{-3mm}
\begin{tabular}{c c c c }
& $E_{L^2}$  & $E_{L^2}$  &
\\
Method &Compton&photoelectric&$D_{\boldsymbol{\chi}}$\\
\hline
\vspace{-2mm}
\\
Proposed method & 0.0908 & 0.1014 & 0.9455 \\
Proposed method without $R_2$ & 0.0818 & 0.1886 & 0.9113 \\
DEFBP & 1.7684 & 293.6438 & 0.0118
\end{tabular}
\label{table:Error40dB}
\end{table}
As observed from Fig. \ref{fig:60dBCorr} and Fig. \ref{fig:60dB} our method provides an accurate reconstruction of the boundaries of the object of interest as well as a reasonable reconstruction of the background images where different objects are clearly visible. On the other hand, DEFBP method provides reasonable reconstructions only in the absence of background noise especially for photoelectric image. It is also observed that using the correlation regularization significantly reduces error in the photoelectric image reconstruction. Quantitative evaluation of the reconstructions in the first example is given in Table \ref{table:Error60dB}. The correlation regularization reduces $E_{L^2}$ for the photoelectric image reconstruction roughly by a factor of three.

\begin{figure*}[t!]
\centering
$\begin{array}{c@{\hspace{0in}}c@{\hspace{0in}}c}

\includegraphics[width=2.1in, trim = 5mm 1mm 5mm 0mm, clip=true]{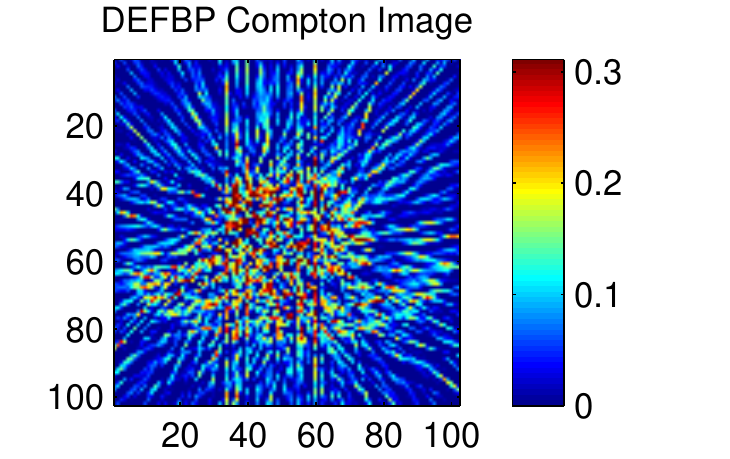}  &
\includegraphics[width=2.1in, trim = 5mm 1mm 5mm 0mm, clip=true]{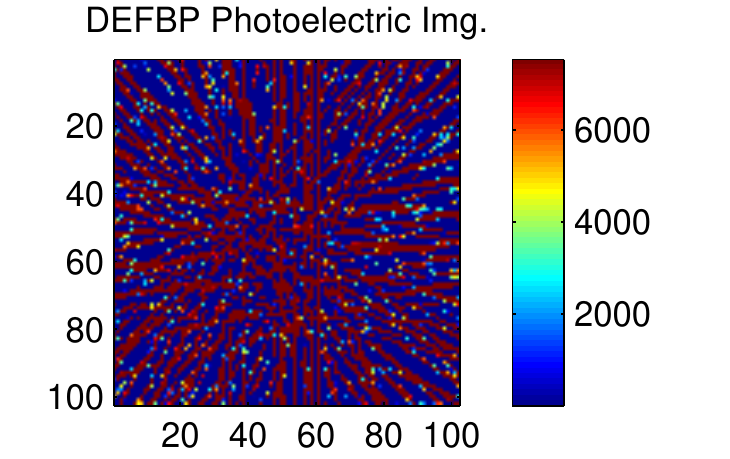} &
\includegraphics[width=1.8in, trim = 14mm 1mm 6mm 0mm, clip=true]{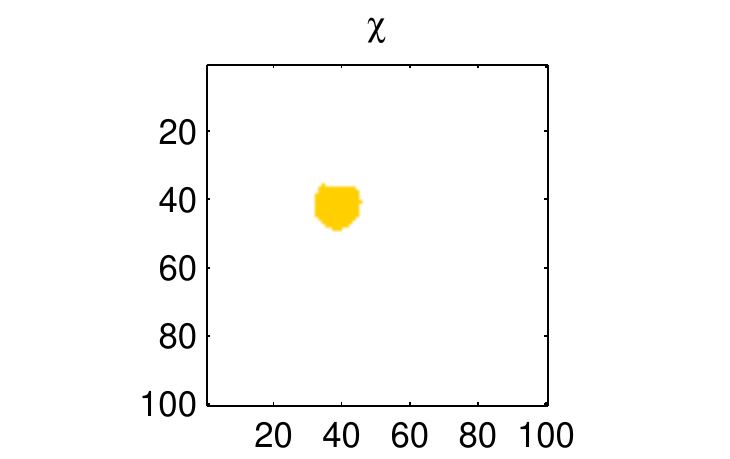}  \\

\includegraphics[width=2.1in, trim = 5mm 1mm 5mm 0mm, clip=true]{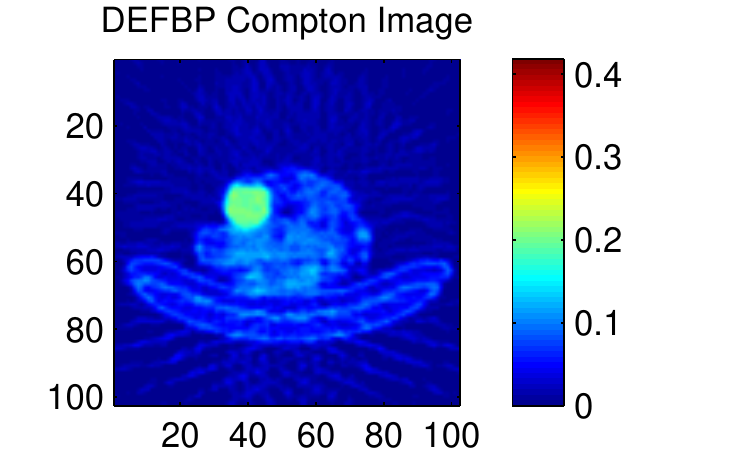}  &
\includegraphics[width=2.1in, trim = 5mm 1mm 5mm 0mm, clip=true]{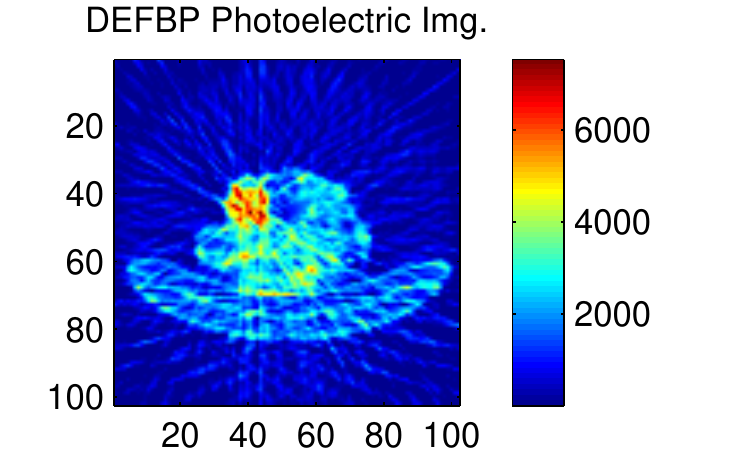} &
\includegraphics[width=1.8in, trim = 14mm 1mm 6mm 0mm, clip=true]{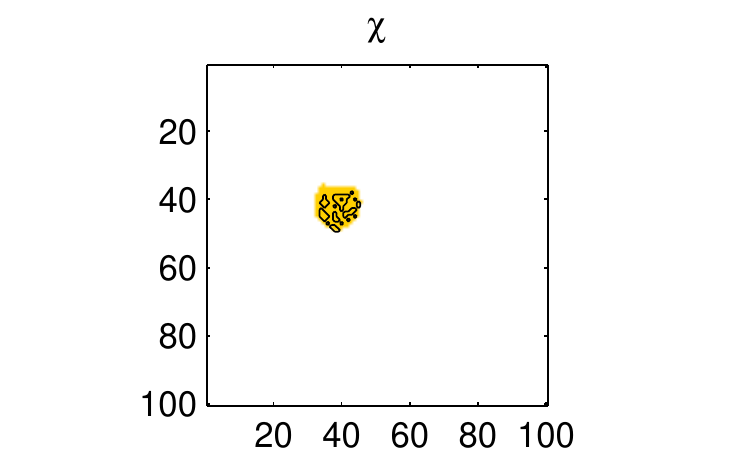}

\end{array}$

\caption{ Simulation with the duffle bag phantom with 40dB background noise. First row: DEFPB reconstruction. Second row: DEFPB reconstruction without background noise. First column: Compton image. Second column: Photoelectric image. Third column: Characteristic function, $\chi$, of the object. The solid yellow is the true object while the thin line represents the estimated boundary.}
\label{fig:DuffleFBP}
\end{figure*}
For the second example the background noise is increased so that SNR was 40dB. Results are demonstrated in Fig. \ref{fig:40dB}. Error in the photoelectric image reconstruction increased compared to the previous example where SNR was 60dB. Nevertheless, the object boundaries are reconstructed quite accurately. Similar to previous case error levels in the reconstructions obtained with DEFBP method are high. Indeed, even for the Compton scatter image, the objects in the scene can hardly be distinguished by eye. See Table \ref{table:Error40dB} for tabulated error values.
\begin{table}
\centering
\caption{Error analysis for the experiment with the duffle bag phantom and 40dB background noise}
\vspace{-3mm}
\begin{tabular}{c c c c}
& $E_{L^2}$  & $E_{L^2}$  &
\\
Method &Compton&photoelectric&$D_{\boldsymbol{\chi}}$\\
\hline
\vspace{-2mm}
\\
Proposed method & 0.1365 & 0.1913 & 0.7468 \\
Proposed method without $R_2$ & 0.1311 & 0.2971 & 0.6325 \\
DEFBP & 1.6702 & 678.99 & 0 \\
DEFBP without b.ground noise.  & 0.1099 & 0.2258 & 0.5026 \\
\end{tabular}
\label{table:ErrorDuffle}
\end{table}

To construct the second phantom we took a DICOM image obtained from a CT scan of a duffel bag and imposed Compton scatter and photoelectric absorption coefficients so that the circular object close to middle was an object of interest and background was composed of low density clutter. The object of interest characteristics are set as the following: $c_0 = 0.3$, $\sigma_c = 0.05$, $p_0 = 5000$, $\sigma_p = 500$. Other parameters of the reconstruction scheme are kept the same as the previous example. We performed the simulation with 40dB background noise. Results are shown in Fig. \ref{fig:Duffle}, Fig. \ref{fig:DuffleFBP} and Table \ref{table:ErrorDuffle}.

\begin{figure*}[t!]
\centering
$\begin{array}{c@{\hspace{0in}}c@{\hspace{0in}}c}

\includegraphics[width=2.1in, trim = 5mm 1mm 5mm 0mm, clip=true]{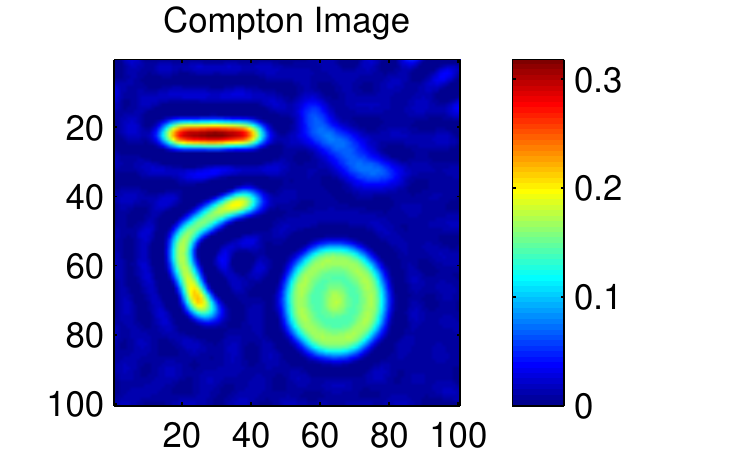}  &
\includegraphics[width=2.1in, trim = 5mm 1mm 5mm 0mm, clip=true]{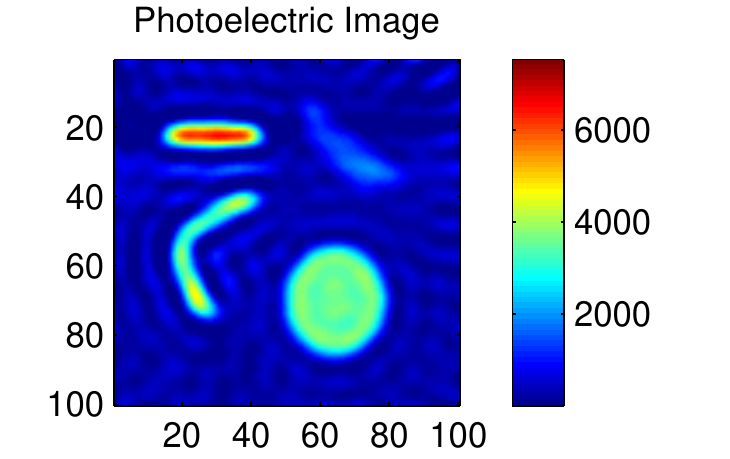} &
\includegraphics[width=1.8in, trim = 14mm 1mm 6mm 0mm, clip=true]{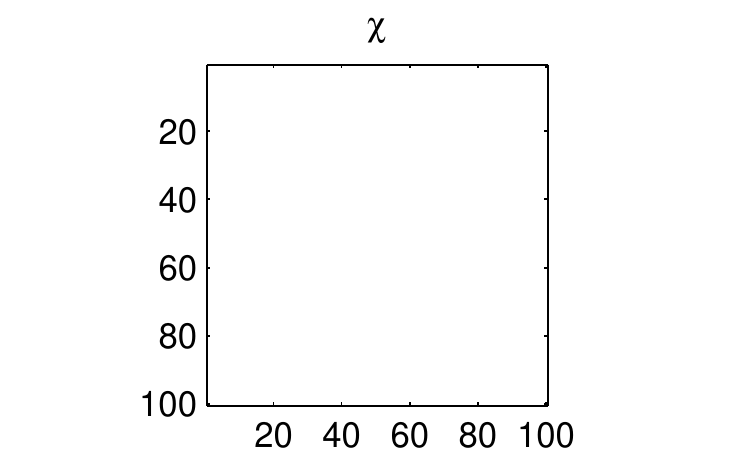}   \\

\end{array}$

\caption{ Simulation with the first phantom with 60dB background noise in the absence of object of interest. First column: Compton image. Second column: Photoelectric image. Third column: Characteristic function, $\chi$, of the object. The solid yellow is the true object while the thin line represents the estimated boundary.}
\label{fig:nobj}
\end{figure*}

In the last example we intend to validate our method in the case where there is no object of interest in the scene. We repeated the first simulation keeping all the parameters but the definition of object of interest characteristics unchanged.  For the definition of $\Gamma$ we used the following values: $c_0 = 0.12$, $\sigma_c = 0.05$, $p_0 = 3000$, $\sigma_p = 500$. The results are shown in Fig. \ref{fig:nobj}. In this example $E_{L^2}$ values for Compton and photoelectric images were $0.1046$ and $0.1498$ respectively. As a result of the absence of an object of interest in the scene the algorithm returns a null characteristic function. The results are demonstrated in Fig. \ref{fig:nobj} and Table \ref{table:nobj}.

\begin{table}
\centering
\caption{Error analysis for the experiment with the absence of object of interest and 60dB background noise}
\vspace{-3mm}
\begin{tabular}{c c c c}
& $E_{L^2}$  & $E_{L^2}$  &
\\
Method &Compton&photoelectric&$D_{\boldsymbol{\chi}}$\\
\hline
\vspace{-2mm}
\\
Proposed method & 0.1046 & 0.1136 & - \\

\end{tabular}
\label{table:nobj}
\end{table}
\section{Conclusions}
\label{sec:conclusions}

A novel semi-shape based polychromatic dual energy CT algorithm with an emphasis on detection of objects whose physical characteristics in terms of Compton scatter and photoelectric absorption coefficients are known to some degree is developed. The \emph{a priori} knowledge about the object characteristics is incorporated into the variational framework via constraints where Levenberg-Marquardt algorithm is used for minimization. Parametric level set approach where the level set function is described via low order basis expansion is implemented. This shape model is capable of representing various kinds of object geometries as shown in Section \ref{sec:numerical_examples}. Along with object of interest boundaries the algorithm provides a reasonable reconstruction of the background images. With the proposed hybrid method we aim to increase the detection rate of objects of interest in an unknown cluttered background. As demonstrated in Section \ref{sec:numerical_examples} accuracy of the photoelectric image reconstruction decreases as the background noise increases; however, object boundaries can still be recovered accurately. Similar to other level set based inverse problem methods, segmentation, hence detection, of the object of interest is simultaneously achieved with reconstruction of background images. This property is advantageous, for example, in airport security applications. Additionally, the use of the proposed correlation based regularization technique improves the reconstruction quality. This regularization technique can be used in inverse problem applications where different parameters, whose sensitivity to measurement data is vastly different, of the same scene needs to be simultaneously reconstructed and spatial similarity between those parameters is expected.

In future work, we intend to test the algorithm in the presence of metals or similar highly attenuating objects that would increase inconsistencies in the measurement data. Low photon count simulations and limited view tomography are also of interest of our research and will be addressed in future work. Another goal is to extend the level set framework so that multiple object types can be assigned as object of interest and treated individually. Additionally, the basis set for the level-set function can be constructed more adaptively. For instance, the widths, the locations and the orientation of each element(blob) in the set can be allowed to change during optimization to achieve greater flexibility. Consideration of different types of basis functions such as compactly supported radial basis functions and wavelets in the representation of the background images is also a promising extension of this work. Regularization parameters are assigned heuristically in this work. For a practically more sound and effective method an automatic determination of regularization parameters is desirable. Finally, in order to make our method applicable to real-sized problems, we will consider increasing the resolution and phantom diameter, therefore optimizing the source code in terms of computational speed-up.

\appendix [Sensitivity Analysis of the Background Reconstruction Problem]

In order to give the reader intuition about the problems in reconstructing the photoelectric absorption, we present a sensitivity analysis of the background estimation problem. We consider a linearized problem and additive Gaussian noise where we analyse the sensitivity of the reconstruction to small chances in the data. We assume no object of interest is present and consider the Jacobian matrices that correspond to background parameters for Compton scatter and photoelectric absorption. For sake of simplicity we further assume that the Jacobian matrix is full rank\footnote{The assumption is reasonable. For instance, in the first example given in Section \ref{sec:numerical_examples}, the size of the Jacobian matrix associated with the data mismatch term is 5640$\times$1354 and its rank was calculated as 1352.} and estimate the errors by the relevant diagonal components of the error covariance matrix of the minimum variance unbiased (MVU) estimator. We provide lower bounds for errors associated with photoelectric and Compton component estimation and show that the bound on the mean square error (MSE) for the latter is significantly smaller.

Assume we have the measurement model for the low energy spectra $S_L(E)$ as in (\ref{projectiondL}). If we plug (\ref{cb}) and (\ref{pb}) into (\ref{projectiondL}) and take the derivatives with respect to $\beta_j$ and $\alpha_j$, we obtain:
\begin{equation}
\label{app:dbeta}
\begin{split}
[\textbf{J}_c]_{ij} & =\frac{\partial [\mathbf{m}_L]_i}{\partial \beta_j } \\
                    & = \frac{[\mathbf{A}]_{i*}[\mathbf{B}]_{*j}}{[\mathbf{m}_L]_i}\int S_L(E)f_{KN}(E)\mathrm{exp}\bigl(-f_{KN}(E)[\mathbf{A}]_{i*}\textbf{B}\boldsymbol{\beta}\\
                    & \quad -f_{p}(E)[\mathbf{A}]_{i*}\textbf{B}\boldsymbol{\alpha}\bigr) \mathrm{d}E
                    \end{split}
\end{equation}
and
\begin{equation}
\label{app:dalpha}
\begin{split}
[\textbf{J}_p]_{ij} & =\frac{\partial [\mathbf{m}_L]_i}{\partial \alpha_j } \\
                    & = \frac{[\mathbf{A}]_{i*}[\mathbf{B}]_{*j}}{[\mathbf{m}_L]_i}\int S_L(E)f_{p}(E)\mathrm{exp}\bigl(-f_{KN}(E)[\mathbf{A}]_{i*}\textbf{B}\boldsymbol{\beta}\\
                    & \quad -f_{p}(E)[\mathbf{A}]_{i*}\textbf{B}\boldsymbol{\alpha}\bigr) \mathrm{d}E.
                    \end{split}
\end{equation}
The spectra $S_L(E)$ consists of $M$ discrete energy levels, therefore we define the matrices $\textbf{W}_\beta$ and $\textbf{W}_\alpha$ with elements
%
%
\begin{equation}
\begin{split}
[\mathbf{W}_{\beta}]_{ij} & = S_L(E_j)\mathrm{exp} \bigl(-f_{KN}(E_j)[\mathbf{A}]_{i*}\textbf{B}\boldsymbol{\beta}\\
                          & \quad -f_{p}(E_j)[\mathbf{A}]_{i*}\textbf{B}\boldsymbol{\alpha}\bigr)
                          \end{split}
\end{equation}
and
\begin{equation}
\begin{split}
[\mathbf{W}_{\alpha}]_{ij} & = S_L(E_j)\mathrm{exp} \bigl(-f_{KN}(E_j)[\mathbf{A}]_{i*}\textbf{B}\boldsymbol{\beta}\\
                           & \quad -f_{p}(E_j)[\mathbf{A}]_{i*}\textbf{B}\boldsymbol{\alpha}\bigr).
                           \end{split}
\end{equation}
Using an appropriate Newton-Cotes numerical integration rule \cite{atkinson2009introduction} with a weight vector  $\boldsymbol{\omega}$ we can rewrite (\ref{app:dbeta}) and (\ref{app:dalpha}) as
\begin{equation}
\textbf{J}_c =\Dm_1\textbf{R}
\end{equation}
and
\begin{equation}
\textbf{J}_p =  \Dm_2\textbf{R}
\end{equation}
where $\Dm_1 =  \mbox{diag}( \textbf{W}_\beta \mbox{diag}(\boldsymbol{\omega}) \textbf{f}_{KN})$, $\Dm_2 =  \mbox{diag}( \textbf{W}_\alpha \mbox{diag}(\boldsymbol{\omega}) \textbf{f}_p)$ and $[\textbf{R}]_{ij} = [\mathbf{m}_L]_i^{-1}[\mathbf{A}]_{i*} [\mathbf{B}]_{*j}$.
But since $\Dm_1$ and $\Dm_2$ are diagonal, hence invertable, we can write
\begin{equation}
\textbf{J}_p = \Dm\textbf{J}_c
\end{equation}
where $\Dm = \Dm_2 \Dm_1^{-1}$.

Now, we can perform a sensitivity analysis with the following linear system of equations
\begin{equation}
\label{appen2}
\Hm f + n = \delta \mathbf{m}
\end{equation}
where $ \Hm = \left[ \textbf{J}_c \quad \Dm \textbf{J}_c \right ] $, $\mathbf{f} = \left[ \boldsymbol{\delta\beta} \quad \boldsymbol{\delta\alpha} \right]^T$. For simplicity here we assume additive white Gaussian noise $\mathbf{n} \sim\mathcal{N}(0,\mathbf{I}\sigma^2) $. With (\ref{appen2}) we investigate how small changes in the measurement data are reflected to the inversion process in the presence of Gaussian noise. If $\Hm \in \mathbb{R}^{M\times N}$, is full rank, the MVU estimator \cite{kay1998fundamentals} is given as
\begin{equation}
\label{appen3}
\hat{\mathbf{f}} = (\Hm^T\Hm)^{-1}\Hm^T\delta \mathbf{m}
\end{equation}
with the associated error covariance matrix, $\mbox{cov}(\mathbf{f}-\hat{\mathbf{f}})=\mbox{E}\{(\mathbf{f}-\hat{\mathbf{f}})(\mathbf{f}-\hat{\mathbf{f}})^T)\}$ given by
\begin{eqnarray}
\label{cov1}
\mbox{cov}(\mathbf{f}-\hat{\mathbf{f})} & = & \sigma^2(\Hm^T\Hm)^{-1} \nonumber \\
&=&\sigma^2 \left[ \begin{array}{cc} \Hm^T\Hm & \Hm^T\Hm \\ \Hm^T\Dm\Hm & \Hm^TD^2\Hm \end{array}
\right ]^{-1} \nonumber \\
&=&\sigma^2 \left[ \begin{array}{cc} \Lam_1^{-1} & \Lam_2^{-1} \\ \Lam_3^{-1} & \Lam_4^{-1} \end{array}
\right ].
\end{eqnarray}
The block matrixes $\Lam_1,\Lam_2,\Lam_3$ and $\Lam_4$ are obtained using the matrix inversion lemma \cite{kaipio2005statistical}. The trace of the covariance matrix can be used to determine the quality of the estimator. Therefore, only $\Lam_1$ and $\Lam_4$  be of interest and each is given as
\begin{equation}
\label{appen4}
\Lam_1 = \Hm^T\Hm-\Hm^T\Dm\Hm(\Hm^T\Dm^2\Hm)^{-1}\Hm^T\Dm\Hm
\end{equation}
and
\begin{equation}
\label{appen6}
\Lam_4 = \Hm^T\Dm^2\Hm-\Hm^T\Dm\Hm(\Hm^T\Hm)^{-1}\Hm^T\Dm\Hm.
\end{equation}

For the estimates $\widehat{\boldsymbol{\delta\beta}}$ and $\widehat{\boldsymbol{\delta\alpha}}$ we have the estimation errors $\mbox{Tr}(\Lam_1^{-1})$ and $\mbox{Tr}(\Lam_4^{-1})$ respectively. Now, the aim is to provide lower bounds for these traces. To start, define $\textbf{M}=\Dm\Hm$  and consider the reduced singular value decomposition (SVD) of  $\textbf{M}$ as $\textbf{M} = \textbf{U}_M\Sigm_M \textbf{V}_M^T$ and of $\Hm$ as $\Hm = \textbf{U}\Sigm \textbf{V}^T$. We can write
\begin{equation}
\label{invGamma1}
\Lam_1^{-1} = \textbf{V}\Sigm^{-1}(\textbf{I}+\textbf{KK}^T)^{-1}\Sigm^{-1}\textbf{V}^T
\end{equation}
and
\begin{equation}
\label{invGamma4}
\Lam_4^{-1} = \textbf{V}_M\Sigm_M^{-1}(\textbf{I}+\textbf{KK}^T)^{-1}\Sigm_M^{-1}\textbf{V}_M^T
\end{equation}
where $\Km=\textbf{U}^T\textbf{U}_M \in \mathbb{R}^{M\times N}$. Since $\textbf{V}$ is orthogonal, $\Lam_1^{-1}$ and $\Sigm^{-1}(I+\Km\Km^T)^{-1}\Sigm^{-1}$ are similar. Likewise, $\Lam_4^{-1}$ and $\Sigm_M^{-1}(\textbf{I}+\Km\Km^T)^{-1}\Sigm_M^{-1}$ are similar. Hence, the traces of $\Lam_1^{-1}$ and $\Lam_4^{-1}$ can be written in terms of these similar matrices as
\begin{equation}
E_{\boldsymbol{\beta}} = \text{tr} \left[ \Sigm^{-1}( \textbf{I} + \Km\Km^T ) \Sigm^{-1} \right]
\end{equation}
and
\begin{equation}
E_{\boldsymbol{\alpha}} = \text{tr} \left[ \Sigm_{M}^{-1}( \textbf{I} + \Km^T\Km ) \Sigm_{M}^{-1} \right].
\end{equation}
Let us define  $m_i = [ \textbf{I} + \Km\Km^T ]_{\mathit{ii}} =  [\textbf{I} + \Km^T\Km ]_{\mathit{ii}}$. Now, we have
\begin{equation}
E_{\boldsymbol{\beta}} = \sum_{i = 1}^{N} \frac{m_i}{\sigma_i^2(\Hm)} \geq \mbox{min}(m_i) \sum_{i = 1}^{N} \frac{1}{\sigma_i^2(\Hm)}
\end{equation}
and
\begin{equation}
\label{e2}
E_{\boldsymbol{\alpha}} = \sum_{i = 1}^{N} \frac{m_i}{\sigma_i^2(\Dm\Hm)} \geq \mbox{min}(m_i) \sum_{i = 1}^{N} \frac{1}{\sigma_i^2(\Dm\Hm)}.
\end{equation}
But, we can write
\begin{equation}
\label{b2}
\begin{split}
\sum_{i = 1}^{N} \frac{1}{\sigma_i^2(\Dm\Hm)} & = \frac {\sum_{i = 1}^{N} \bigl(\prod_{k\neq i}\sigma_k^2(\Dm\Hm)\bigr)}{\prod\sigma_k^2(\Dm\Hm)} \\
                                              & = \frac{\sigma_1^2(\Dm\Hm) }{\sigma_1^2(\Dm\Hm)} \frac {\sum_{i = 1}^{N} \bigl(\prod_{k\neq i}\sigma_k^2(\Dm\Hm)\bigr)}{\prod\sigma_k^2(\Dm\Hm)} \nonumber \\
& \geq \frac{N \prod \sigma_k^2(\Dm\Hm)}{\sigma_1^2(\Dm\Hm) \prod \sigma_k^2(\Dm\Hm) } = \frac{N }{\sigma_1^2(\Dm\Hm)}
\end{split}
\end{equation}
here $\sigma_1(\Dm\Hm)$ is the largest singular value of $\Dm\Hm$. If we plug (\ref{b2}) into (\ref{e2}) we get
\begin{equation}
\label{e22}
E_{\boldsymbol{\alpha}} \geq \frac{N \mbox{min}(m_i) }{\sigma_1^2(\Dm\Hm)}.
\end{equation}
Similarly, for $E_{\boldsymbol{\beta}}$ we have
\begin{equation}
\label{e11}
E_{\boldsymbol{\beta}} \geq \frac{N \mbox{min}(m_i)}{\sigma_1^2(\Hm)}.
\end{equation}
Next, we use the following inequality \cite{horn1994topics} for the largest singular value of a matrix $\Hm$
\begin{equation}
\label{eig_bound}
\sigma_1(\Hm)  \leq \left[ \Vert \Hm \Vert_1 \Vert \Hm \Vert_{\infty} \right]^{1/2}.
\end{equation}
Using (\ref{eig_bound}) we can write
\begin{equation}
\sigma_1(\Dm\Hm)\leq \Vert \Dm\Hm \Vert_1 \Vert \Dm\Hm \Vert_{\infty} \leq
d_{\mbox{max}}^2\left( \Vert \Hm \Vert_1 \Vert \Hm \Vert_{\infty} \right)
\end{equation}
where, $d_{\mbox{max}}$ is $\mbox{max}\left([D]_{ii}\right)$. Now, (\ref{e11}) and (\ref{e22}) becomes
\begin{equation}
\label{e222}
E_{\boldsymbol{\alpha}} \geq \frac{N \mbox{min}(m_i) }{d_{\mbox{max}}^2\left( \Vert \Hm \Vert_1 \Vert \Hm \Vert_{\infty} \right)}
\end{equation}
and
\begin{equation}
\label{e111}
E_{\boldsymbol{\beta}} \geq \frac{N \mbox{min}(m_i)}{\Vert \Hm \Vert_1 \Vert \Hm \Vert_{\infty} }.
\end{equation}%

The lower bound of $E_{\boldsymbol{\alpha}}$ is $d_{\mbox{max}}^{-2}$ times bigger than that of $E_{\boldsymbol{\beta}}$. Our aim is to compare these lower bounds; hence at this point, we can be more concrete. For instance, consider a homogeneous $20\times 20$ cm area illuminated from angle $\phi=0$ by a parallel beam of X-rays using the low energy spectra $S_L(E)$. For this set up, we performed calculations corresponding three different scenario where the medium is filled with water, plexiglass and aluminium. Photoelectric absorption and Compton scatter coefficients \cite{de2001iterative} of these materials and calculated $d_{\mbox{max}}^{-2}$ values are given in Table \ref{table:sensitivity}. For instance, in case the medium is water the value of $d^{-2}$ is $4.3841\times10^{5}$ which is approximately $15$ times bigger than than the ratio of the actual photoelectric absorption and Compton scatter coefficients. Also note that the difference between the lower bounds increases dramatically as the attenuation characteristics of the medium increases as in the case of aluminium.
\begin{table}
\centering
\label{table:sensitivity}
\caption{Sensitivity Analysis}
\begin{tabular}{|c|c|c|c|}
\hline
Material &  $p$   &   $c$    &   $d_{\mbox{max}}^{-2}$ \\
\hline
Water    &  4939.2 &  0.1907 &  $4.3841\times 10^{5}$ \\
\hline
Plexiglass    &  3670.1 &  0.2157  &   $4.3841\times 10^{5}$ \\
\hline
Aluminium    &  72887.5 &  0.4547 &   $7.7815\times 10^{6}$ \\
\hline
\end{tabular}
\end{table}

The lower bound for the error made while estimating the perturbations in photoelectric absorption coefficients is significantly bigger than the error associated with the estimation of perturbations in Compton scatter coefficients. From this point of view, we expect a similar behavior in the original (nonlinear) problem especially since Gauss-Newton (e.g., Levenberg-Marquardt) type algorithms rely on linearization of the problem at the current estimate to find the next estimate at each iteration. We can, therefore, conclude  that photoelectric image reconstruction is more prone to error compared Compton image reconstruction. In this work we tried to address this issue using the regularization term given in (\ref{R2}).

\section*{Acknowledgment}

This work supported by the U.S. Department of Homeland Security under Award Number 2008-ST-061-ED0001.  The views and conclusions contained in this document are those of the authors and should not be interpreted as necessarily representing the official policies, either expressed or implied, of the U.S. Department of Homeland Security.

The authors wish to thank Prof. Misha Kilmer for her contributions to the analysis in the Appendix.

\ifCLASSOPTIONcaptionsoff
  \newpage
\fi

\end{document}